\documentclass[10pt,twocolumn]{article}

\usepackage[margin=0.75in]{geometry}
\usepackage[T1]{fontenc}
\usepackage{lmodern}
\usepackage{booktabs}
\usepackage{graphicx}
\usepackage{amsmath}
\usepackage{array}
\usepackage{caption}
\usepackage{float}
\usepackage{stfloats}   
\fnbelowfloat           
\usepackage{enumitem}
\usepackage{multirow}
\usepackage{titlesec}
\usepackage{balance}
\usepackage[section]{placeins}  
\usepackage{xcolor}
\usepackage{tikz}
\usetikzlibrary{arrows.meta,positioning}
\usepackage[protrusion=true,expansion=false]{microtype}
\usepackage[hidelinks]{hyperref}
\usepackage[all]{hypcap}
\usepackage{listings}
\lstdefinestyle{promptstyle}{
  basicstyle=\scriptsize\ttfamily,
  breaklines=true,
  breakatwhitespace=false,
  columns=fullflexible,
  keepspaces=true,
  showstringspaces=false,
  frame=single,
  framesep=4pt,
  xleftmargin=4pt,
  xrightmargin=4pt,
  aboveskip=6pt,
  belowskip=4pt,
}

\hypersetup{
  pdftitle={Inject or Navigate? Token-Efficient Retrieval for LLM Analysis of Transactional Legal Documents},
  pdfauthor={Mahmoud Hany, Mourad ElSheraey, Mahmoud Said, Peter Naoum}
}

\titleformat{\section}{\large\bfseries\raggedright}{\thesection}{0.6em}{}
\titleformat{\subsection}{\normalsize\bfseries\raggedright}{\thesubsection}{0.6em}{}
\titlespacing*{\section}{0pt}{1.6ex plus 1ex}{0.9ex}
\titlespacing*{\subsection}{0pt}{1.2ex plus 0.8ex}{0.6ex}
\setlist{itemsep=2pt,topsep=3pt,leftmargin=1.4em}
\renewcommand{\arraystretch}{1.08}

\newcommand{\inject}{\textsc{inject}}
\newcommand{\navigate}{\textsc{navigate}}
\newcommand{\navembed}{\textsc{nav\-embed}}
\newcommand{\navindex}{\textsc{nav\-index}}

\begin{document}

\twocolumn[{%
\begin{center}
{\LARGE\bfseries Inject or Navigate? Token-Efficient Retrieval for\\[2pt]
LLM Analysis of Transactional Legal Documents}\\[1.0em]
{\small
\begin{tabular}{cccc}
Mahmoud Hany\textsuperscript{*} & Mourad ElSheraey\textsuperscript{*} & Mahmoud Said\textsuperscript{*} & Peter Naoum \\[0.15em]
{\footnotesize\texttt{hany@syntheia.app}} & {\footnotesize\texttt{mourad@syntheia.io}} & {\footnotesize\texttt{mahmoud@syntheia.app}} & {\footnotesize\texttt{peter@syntheia.app}} \\
\end{tabular}}\\[0.6em]
{\footnotesize Syntheia Pty Ltd}
\end{center}
\vspace{0.8em}
\begin{center}
\begin{minipage}{0.9\textwidth}
\small
\noindent\textbf{Abstract.}\;
Answering questions over a set of transactional legal documents is most simply
done by \emph{injecting} the whole corpus into the LLM's context window on
every query. That baseline maximises retrieval recall, but its token footprint
scales with the corpus rather than the question, and long-context degradation
scales with it. We report what it took to replace full-corpus injection in a
legal-document analysis system, comparing it against two structured retrieval
modes over our proprietary structure-aware chunking: embedding retrieval
(\navembed{}) and LLM navigation over a compact structured index
(\navindex{}). On a 20-question benchmark with verified ground-truth answers, a
position-bias-controlled, reference-anchored pairwise judge scored
semantic retrieval with reranking tied with injection on 16 of 18
document-bound questions (injection preferred on 2) while attending to
$17.3\times$ fewer input tokens (a general-text-embedding (GTE) configuration
reaches $29.9\times$ at a lower tie rate); both modes were judged tied on the 2 out-of-scope
controls. \navindex{} was judged tied on all 18 at a
$1.61\times$ smaller total token footprint, a ${\sim}56\times$ smaller
answering context, and 25\% lower dollar cost. We derive a closed-form
caching-crossover rule: cached injection is cheaper in dollars only while the
corpus stays below roughly ten times the retrieval payload. Scope and uncertainty are quantified in
Section~\ref{sec:threats}.
\vspace{1.2em}
\end{minipage}
\end{center}
}]

{\renewcommand{\thefootnote}{\fnsymbol{footnote}}%
\footnotetext[1]{Equal contribution. Mahmoud Hany built and ran the
evaluation program across all three phases and wrote the paper's initial
draft. Mourad ElSheraey revised the analysis, added the uncertainty
quantification and threats-to-validity treatment, and prepared the final
manuscript. Mahmoud Said implemented the cross-encoder reranking pipeline and
metadata fields, and contributed the cross-reference extraction. Peter Naoum
implemented the initial codebase, and supervised the experimental work.}}

\section{Introduction}
\label{sec:intro}

Large language models have made it practical to interrogate documents directly:
a user poses a question in natural language and receives an answer grounded in the
source text, with no hand-built query interface and no manual review. Few
professions stand to benefit more from this than law, whose day-to-day work consists
largely of locating, reconciling, and interpreting provisions spread across long,
densely cross-referenced instruments. Syntheia provides the data layer for
this workload: a structured document index over coordinated
\emph{transaction sets} (an original agreement plus its amendments, side
letters, schedules, and disclosure documents) served through an API, and
directly to LLMs such as Claude via the Model Context
Protocol~\cite{anthropic2024mcp}, for document
tools to build on. This paper reports the evaluation that decided how
retrieval over that index should work: what it took to replace full-corpus
injection with structured retrieval without giving up answer quality, and
what each alternative costs in tokens and dollars.

The simplest method is to place the entire corpus in the model's context window on
every query. We call this approach \inject{}, and we adopt it as our baseline
deliberately, because it represents the upper bound on retrieval recall. Every clause,
defined term, and cross-reference is in context on every query, so there is nothing
for a retriever to miss and nothing to misconfigure. An alternative that a
careful judge cannot distinguish from \inject{} is therefore a stronger result
than one that merely outperforms a lossy retriever; the corpus was
deliberately sized to keep \inject{} runnable (Section~\ref{sec:method}).

Two costs limit \inject{} as the corpus grows. The token footprint scales with the
corpus rather than the question (every query pays for every clause regardless of
relevance) and the corpus cannot exceed the context window at all. Separately, LLMs
attend unevenly across long inputs and lose information positioned in the middle of
the context~\cite{liu2024lost}. The legal domain compounds both: answers routinely
depend on defined terms, on cross-references between schedules and operative
clauses, and on how an original agreement interacts with later
amendments, so the text bearing on a question is dispersed
across the corpus and easily diluted by surrounding material. Existing contract
benchmarks (CUAD~\cite{hendrycks2021cuad}, ContractNLI~\cite{koreeda2021contractnli})
operate on single documents and do not capture this cross-instrument
dependency, which remains an open gap for legal retrieval evaluation.

Both modes studied here operate over the same input representation: documents
chunked by a proprietary structure-aware method developed at Syntheia, rather
than split into fixed-size spans. Chunking is held constant as a fixed
preprocessing step, so that what we measure is the retrieval strategy and not
the chunking; we do not evaluate naive fixed-size chunking, whose tendency to
fragment provisions and reduce retrieval quality is already well
documented~\cite{merola2025chunking,taiwo2026chunking}. We report two
comparisons:

\smallskip
\noindent\textbf{Comparison~1} (Section~\ref{sec:embed}) evaluates
\emph{Navigate-Embeddings} (\navembed{}), which pre-embeds the document nodes and,
at query time, retrieves the top-$k$ most similar by cosine similarity, placing only
those nodes in the answering model's context. Twelve configurations (RUN-001--012)
were run in three phases: Phases~I--II while the harness matured
(RUN-001--008), Phase~III at the full 20-question scale (RUN-009--012). Each is an internally
controlled paired comparison (\inject{} versus \navembed{} on identical
questions), and the headline claims rest on the Phase~III runs.

\smallskip
\noindent\textbf{Comparison~2} (Section~\ref{sec:json}) evaluates
\emph{Navigate-Index} (\navindex{}), which replaces vector similarity with guided
navigation over a compact structured index: boolean semantic flags, generated
summaries, and explicit cross-reference and defined-term graphs. We sweep six
descriptor configurations (the per-node metadata the retriever sees) to
characterise the quality--token-footprint trade-off.

\smallskip
The two \navigate{} modes reduce \inject{}'s burden along two dimensions that
prompt caching has pulled apart: token footprint and dollar cost. Caching
lowers the \emph{price} of re-reading a large context without reducing the
tokens the model must attend over when it answers, so the two dimensions are
tracked separately throughout (Section~\ref{sec:tokens-dollars}). The cache
discount also holds only within a single warm session (a run of queries close
enough in time to keep the provider's short-lived prompt cache live), and a
corpus that outgrows the context window cannot be injected at any price
(Section~\ref{sec:crossover}).

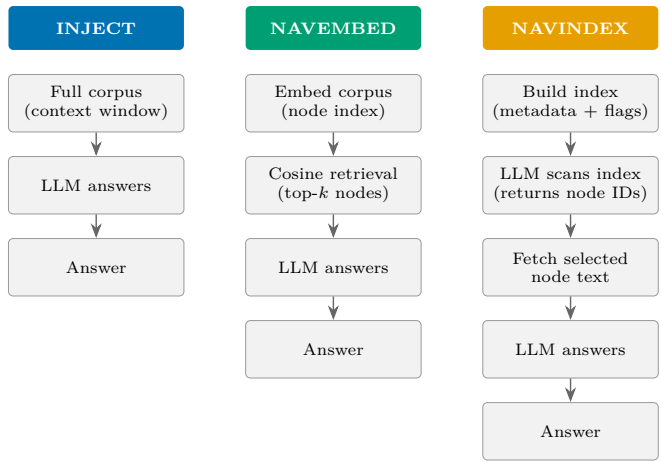
\begin{figure}[htbp]
\centering
\resizebox{\linewidth}{!}{%
\begin{tikzpicture}[
  font=\scriptsize,
  box/.style={draw=black!40, rounded corners=2pt, fill=black!5, align=center,
              minimum width=26mm, minimum height=8.5mm, inner sep=2.5pt},
  hdr/.style={rounded corners=2pt, align=center, minimum width=26mm,
              minimum height=6.5mm, text=white, font=\scriptsize\bfseries},
  arr/.style={-{Stealth[length=2.2mm]}, black!60, semithick},
  node distance=3.5mm
]
\node[hdr, fill={rgb,255:red,0;green,114;blue,178}] (i0) {INJECT};
\node[box, below=of i0] (i1) {Full corpus\\(context window)};
\node[box, below=of i1] (i2) {LLM answers};
\node[box, below=of i2] (i3) {Answer};
\draw[arr] (i1) -- (i2);
\draw[arr] (i2) -- (i3);
\node[hdr, fill={rgb,255:red,0;green,158;blue,115}, right=9mm of i0] (e0) {NAVEMBED};
\node[box, below=of e0] (e1) {Embed corpus\\(node index)};
\node[box, below=of e1] (e2) {Cosine retrieval\\(top-$k$ nodes)};
\node[box, below=of e2] (e3) {LLM answers};
\node[box, below=of e3] (e4) {Answer};
\draw[arr] (e1) -- (e2);
\draw[arr] (e2) -- (e3);
\draw[arr] (e3) -- (e4);
\node[hdr, fill={rgb,255:red,230;green,159;blue,0}, right=9mm of e0] (n0) {NAVINDEX};
\node[box, below=of n0] (n1) {Build index\\(metadata $+$ flags)};
\node[box, below=of n1] (n2) {LLM scans index\\(returns node IDs)};
\node[box, below=of n2] (n3) {Fetch selected\\node text};
\node[box, below=of n3] (n4) {LLM answers};
\node[box, below=of n4] (n5) {Answer};
\draw[arr] (n1) -- (n2);
\draw[arr] (n2) -- (n3);
\draw[arr] (n3) -- (n4);
\draw[arr] (n4) -- (n5);
\end{tikzpicture}}
\caption{The three pipeline modes. \textbf{Left:} \inject{}: full corpus in context on every query. \textbf{Centre:} \navembed{}: cosine retrieval selects the top-$k$ nodes at query time. \textbf{Right:} \navindex{}: three-step structured navigation over a compact boolean-flagged index, no embedding similarity required.}
\label{fig:arch}
\end{figure}

Within that frame, the two modes sit at different points on the
quality--token-footprint curve (Table~\ref{tab:summary}): \navindex{}
maximises structural fidelity at a modest token-footprint gain, \navembed{}
maximises token-footprint reduction, and the dollar comparison between them and cached
\inject{} is corpus-dependent (Section~\ref{sec:crossover}).
Figure~\ref{fig:arch} illustrates all three modes.

The contributions are: (i)~a paired evaluation of full-corpus injection against
two structured retrieval modes on a scored legal-QA evaluation set with verified
ground-truth answers, judged by a position-bias-controlled, reference-anchored
pairwise protocol (answer-level equivalence; retrieval recall is not directly
measured), with uncertainty quantified throughout; (ii)~the design of the
\navindex{} format, a navigation descriptor for legal instruments (typed
boolean flags, cross-reference and defined-term graphs, a hard selection
cap),
documented to field level by the released prompts, which also operationalise
amendment precedence and defined-term resolution
(Appendix~\ref{app:prompts}); (iii)~a closed-form
caching-crossover rule separating token footprint from dollar cost, with its
session-economics caveats; and (iv)~a catalogue of failure modes we would not
ship (BM25 length bias, titles-only over-selection, uncapped node
selection) and an account of the evaluation's limits.

\section{Related Work}
\label{sec:related}

RAG couples a parametric generator with a non-parametric retriever so that
generation is conditioned on fetched evidence~\cite{lewis2020rag}; recent surveys
map the design space~\cite{gao2023survey}. Dense passage
retrieval~\cite{karpukhin2020dpr} established learned bi-encoders as a strong
default, followed by a family of general-purpose sentence embeddings
(Sentence-BERT~\cite{reimers2019sbert}, E5~\cite{wang2022e5}, GTE~\cite{li2023gte},
BGE~\cite{xiao2023bge}, and MPNet-based encoders~\cite{song2020mpnet}). BM25 remains a strong lexical
baseline~\cite{robertson2009bm25}; hybrid systems fuse dense and sparse signals via
Reciprocal Rank Fusion~\cite{cormack2009rrf}, and cross-encoder
rerankers~\cite{nogueira2019rerank} re-score candidates at higher fidelity. Our
strongest \navembed{} configuration is a retrieve-then-rerank pipeline; our hybrid
BM25 run illustrates a known failure mode (length bias) on clause-structured text.
Chunking granularity affects retrieval
quality~\cite{merola2025chunking,taiwo2026chunking}; we hold structure-aware
chunking fixed and vary only the retrieval strategy.

Two observations make token footprint a first-class metric here. Long-context
performance degrades when relevant information sits in the middle of a long
input~\cite{liu2024lost}, and provider-side prompt caching discounts re-reads of a
fixed prefix~\cite{anthropic2024caching}, lowering the \emph{price} of a large
context without lowering the \emph{tokens} the model attends to. Dollar cost and
token footprint must therefore be tracked as separate quantities. For evaluation,
LLM-as-judge protocols are common~\cite{zheng2023judge} but exhibit position
bias~\cite{wang2023fair}; our forward/reverse protocol is a direct response.

In legal NLP, CUAD~\cite{hendrycks2021cuad} and LEDGAR~\cite{tuggener2020ledgar}
are the leading contract benchmarks, and ContractNLI~\cite{koreeda2021contractnli}
shows that entailment over contracts requires resolving defined terms and
cross-references, structural dependencies that flat retrieval cannot
represent. Hierarchical indexing organises nodes into a tree and navigates from
coarse summaries to fine provisions~\cite{sarthi2024raptor}; a complementary line
uses typed metadata pre-filters to narrow an index before any similarity
computation, analogous to predicate push-down in query optimisation. Legal
documents require both: defined terms propagate meaning across sections,
cross-references bind operative clauses to schedules, and amendments supersede
base provisions in ways neither cosine similarity nor BM25 can represent. The
\navindex{} format encodes these signals directly and sits in the
hierarchical-navigation paradigm rather than in flat RAG; among vectorless
navigators it is closest to PageIndex~\cite{pageindex2025}
(Section~\ref{sec:descriptor}).

\section{Problem Setting and Method}
\label{sec:method}

\subsection{Corpus and models}

\textbf{Document corpus.}
The benchmark indexes fifteen transactional legal agreements drawn from three
practice areas. Six \emph{credit facility agreements} (\texttt{FA}) cover two
borrowers: Fluor Corporation (a \$1.8\,B revolving facility plus one amendment)
and Generac (an ABL facility plus two term-loan amendments and a 2024
refinancing). Three \emph{limited partnership agreements} (\texttt{LP}) cover two
funds: Thomas Green Fund, a US real-estate fund targeting LEED-certified office
and multi-family assets with a \$500\,M commitment cap and a companion side
letter, and Carlyle PE Partners, a PE fund with a 5\,\% IRR hurdle and 12.5\,\%
incentive allocation. Six \emph{share/asset purchase agreements and disclosure
schedules} (\texttt{SP}) cover four transactions: Meta's divestiture of Giphy to
Shutterstock (\$128\,M purchase price), the First Avenue Networks disclosure
schedule, the PPG/Comex cross-border acquisition, and a Royal Wolf Holdings
facility agreement. Two additional Netflix facility agreements are indexed in the
open-search workspace but not bound to any scored question. The scored benchmark
runs use six of the fifteen documents: one Thomas Green Fund LPA, one Carlyle PE
Partners LPA, the Meta/Giphy SPA, the First Avenue Networks disclosure schedule,
the Generac 2024 amendment, and the Fluor facility amendment.
All fifteen documents are public filings drawn from SEC EDGAR exhibits,
Kentucky pension FOIA releases, and UK government sources. The scored question
set with its verified reference answers and a source reference for every
document is released as supplementary material; all prompts are printed in
Appendix~\ref{app:prompts}.

\textbf{Question set.}
The \emph{scored evaluation set} contains 20 questions with verified
ground-truth answers (Appendix~\ref{app:questions}): 18 are bound to specific
source documents (six per practice area, three questions per document), and 2
are \emph{out-of-scope controls} whose answers are deliberately absent from all
indexed documents, testing whether the system correctly declines rather than
hallucinating. Because the controls test declination rather than retrieval, we
report them separately from the document-bound counts throughout. The
first two phases ran on a 15-question, document-pinned subset of
\emph{DocNavBench}, a companion 52-question open-search set over the same
corpus, authored by the reviewing lawyer: its questions are deliberately
issue-driven, written as a practitioner would ask \emph{before} reading and
without presuming the answer exists (so ``that is not in the agreement'' can
be the correct finding), and it is document-agnostic, pairwise-judged only,
with no ground-truth answers. The Phase~I--II runs used 3 questions first,
widening to 9 and then all 15 as the harness matured (Appendix~\ref{app:dev});
the \navindex{} descriptor sweep (Configs~1--5, Section~\ref{sec:json}) also
belongs to Phase~II, sharing its pool and corpus. The 20-question scored
set was written fresh at Phase~III by the engineering team: questions and
reference answers were drafted with the same model family over the six scored
documents, verified against the sources, and endorsed by the reviewing
lawyer, with the out-of-scope controls added so that every document-bound
question is answerable from the indexed documents by construction. It shares
no question with the Phase~I--II pool, so nothing the pipeline was tuned
on is scored. The lawyer's assessment of the set's difficulty and phrasing
is reported in Section~\ref{sec:threats}. The four Phase~III runs and
\navindex{} Config~6 all use the scored set; DocNavBench's full open-search
evaluation is future work. The answering model is
\texttt{claude-sonnet-4-6}; the judge is \texttt{claude-opus-4-7}.

\textbf{Context-window scope.}
The scored evaluation was deliberately sized to stay within the model's context
window, so that \inject{} is a runnable baseline and the comparison is like-for-like
on every question. This is a constraint of the evaluation, not of the methods. Real
legal matters frequently exceed the window (a single transaction can span dozens of
instruments across base agreements, amendments, side letters, and schedules) and
beyond that point \inject{} is not a viable method rather than merely an expensive
one, so it would not appear as a baseline at all. The \navigate{} modes carry only a
bounded retrieval payload into the answering context and are not subject to this
ceiling. Within-window parity is thus the deliberately hard test: a mode that
cannot match \inject{} where \inject{} is runnable could not be trusted where
\inject{} is not.

\subsection{Judging protocol}
\label{sec:judging}

Every question is answered \emph{twice, in parallel}: once by \inject{} and once
by the \navigate{} variant under test. The judge receives the question, the
\emph{verified reference answer}, and both model answers, and decides which
answer more accurately reflects the reference, or a tie if both are equally
correct or equally wrong (the full judge prompt is Appendix~\ref{app:prompts},
\texttt{JUDGE\_GROUNDTRUTH\_SYSTEM}). Each pair is judged in two directions,
\emph{forward} and \emph{reverse} (positions swapped); a \emph{win} is awarded
only when both directions agree, and a hedged verdict in either direction
(e.g.\ ``slight lean toward A, effectively tied'') is treated as a tie. The
two-direction rule controls the position bias documented for LLM
judges~\cite{wang2023fair,zheng2023judge}.

A tie under this protocol therefore certifies that the two answers are
\emph{judged equally consistent with the verified reference}, equally
correct or equally wrong relative to ground truth, and not merely that they
read alike. It does not certify that either answer is right; judge error is bounded
empirically by the audit below, and the remaining caveats (a forced-decision probe, i.e.\ a rerun with ties
disallowed; judge--answerer provider overlap; and tie-category granularity)
are consolidated in Section~\ref{sec:threats}. All 20-question scored runs
(RUN-009--012 and \navindex{} Config~6) were judged with this reference-anchored
prompt; the Phase~I--II runs, \navindex{} Configs~1--5 included, were scored
with the Phase~I--II preference judge, which compares the two answers without
a reference (reproduced in Appendix~\ref{app:prompts}).

\textbf{Judge audit.} A manual audit of one 15-question run classified $9/15$
verdicts as clearly correct, $4/15$ as defensible-but-debatable, and $2/15$
($13.3\%$) as likely wrong (95\% Clopper--Pearson interval~\cite{clopper1934} $[2\%, 40\%]$).
The audit was performed by
one author against the verified ground-truth answers for that run; a verdict
was ``clearly correct'' if the judge's preferred answer matched the ground
truth exactly, ``defensible-but-debatable'' if partially correct or a
reasonable legal interpretation, and ``likely wrong'' if it contradicted the
ground truth. No second annotator reviewed the run, so inter-annotator
agreement is not available.

\subsection{Token footprint versus dollar cost}
\label{sec:tokens-dollars}

Prompt caching separates two axes. \emph{Token footprint} is every
input token the model attends to, including cached context; it governs the
long-context degradation noted above and the hard window limit.
\emph{Dollar cost} is recorded operational spend,
reshaped by discounted cache reads and one-time cache
writes~\cite{anthropic2024caching}. Caching can make a full-context \inject{}
cheap to re-read without reducing the tokens inside the model's attention. In
RUN-005 (15 questions), \inject{} attended to $3{,}217{,}792$ input tokens in
total (${\sim}214.5$K per question) while \navembed{} processed $171{,}307$:
about $3\times$ apart on dollars (\$1.881 vs.\ \$0.611) but $18.8\times$
apart on token footprint. Caching compresses the dollar axis; the
token-footprint gap is what remains structural.

\subsection{System context}
\label{sec:system}

The three modes differ in their runtime shape as well as their token profile.
\inject{} issues one call per question over a long (cache-discounted) prefix.
\navembed{} embeds the query, retrieves and optionally reranks nodes, then
issues one short answering call. \navindex{} issues two sequential LLM calls per
question (a short-output index scan over a cached prefix, then a short
answering call), trading tokens for one extra round-trip of latency.
Indexing is a one-time, per-document cost with two layers: headings, clause
references, and the cross-reference and defined-term edges are extracted
deterministically by the chunking pipeline (same input, same output,
verifiable against the source), while the semantic boolean flags
(Table~\ref{tab:flags}), keyword tags, and per-node summaries are populated
by a separate enrichment pass of the Syntheia indexing pipeline.

\section{INJECT vs.\ NAVEMBED}
\label{sec:embed}

\subsection{Approach}

\navembed{} pre-indexes each document into a tree of \emph{nodes}. Every node
carries the section's verbatim text plus a descriptor used at index time.
At query time the query is embedded with the same model and cosine similarity
selects the top-$k$ nodes; only those nodes (plus the question) are placed in
the answering model's context. This replaces the full corpus with a small,
question-relevant working set. Five retrieval strategies were evaluated:

\begin{enumerate}[label=\textbf{R\arabic*.},leftmargin=2.0em,itemsep=2pt]
\item \textbf{Embedding top-$k$}~\cite{karpukhin2020dpr}\textbf{:} cosine
  similarity; keep the top $k$.
\item \textbf{Embedding $+$ cross-reference expansion:} after top-$k$ hits,
  follow each hit's \texttt{crossReferencedIds} and add those nodes.
\item \textbf{Hybrid semantic $+$ BM25:} fuse dense and lexical
  signals~\cite{robertson2009bm25,cormack2009rrf}.
\item \textbf{Semantic $+$ rerank:} retrieve $40$ candidates by cosine,
  rerank with a cross-encoder, keep the top $10$~\cite{nogueira2019rerank}.
\item \textbf{Naive LLM index scan:} the model sees a titles-only index,
  selects candidates, and those are fetched. This prototype over-selected
  ($87$, $72$, and $38$ nodes on its three trial questions, pushing the
  token footprint to within $1.2\times$ of \inject{}) and motivated the
  \navindex{} format described in the next section.
\end{enumerate}

Twelve configurations were run in three phases; the earlier two
(3--15 questions, evolving harness) are reported as run history in
Appendix~\ref{app:dev}, and all claims below rest on the Phase~III runs, which
share the full 20-question set, the same six-document corpus, and the same
prompts.

\subsection{Results}
\label{sec:embed-results}

\begin{table}[!b]
\centering
\caption{Phase~III \navembed{} runs (20 questions: 18 document-bound $+$ 2
out-of-scope controls). ``Red.''\ (reduction) is the \inject{}/\navembed{} input-token
ratio; dollar columns are total recorded run cost in USD
(Section~\ref{sec:crossover} price card). Judge outcomes are
ties\,/\,\inject{} wins\,/\,\navembed{} wins.
RUN-010 uses ground-truth node labels to validate retrieval and is an oracle
upper bound, not a deployable configuration. Per-question verdicts are printed
for RUN-012 in Appendix~\ref{app:detail}; its two \inject{} wins are
document-bound and both controls were judged tied. RUN-009's \inject{} arm
ran fresh (\$1.621); RUN-010--012 reuse a cached \inject{} answer set
(recorded at \$1.614), so their cross-run variation is retrieval-side only.}
\label{tab:main}
\footnotesize
\setlength{\tabcolsep}{3.2pt}
\begin{tabular}{llcccc}
\toprule
Run & Method & T/I/N & Red. & Nav.\ \$ & Inj.\ \$ \\
\midrule
RUN-009 & GTE embed          & 15/3/2 & \textbf{29.9$\times$} & 0.406 & 1.621 \\
RUN-011 & Hybrid $+$ BM25    & 16/4/0 & 8.5$\times$  & 1.217 & 1.614 \\
RUN-012 & Semantic $+$ rerank & \textbf{18/2/0} & 17.3$\times$ & 0.644 & 1.614 \\
\midrule
\multicolumn{2}{l}{RUN-010 (oracle; excluded)} & 18/2/0 & 18.7$\times$ & 0.601 & 1.614 \\
\bottomrule
\end{tabular}
\end{table}

Table~\ref{tab:main} reports the Phase~III runs; Table~\ref{tab:stats} gives
exact 95\% confidence intervals for every headline rate, and
Appendix~\ref{app:dev} reports the full twelve-run history
(Table~\ref{tab:glance}), across which the token-footprint reduction held at
$8.5$--$29.9\times$.

The strongest deployable result is semantic retrieval with reranking
(RUN-012)\footnote{The initial 40-candidate retrieval uses the
\texttt{paraphrase-\allowbreak MiniLM-\allowbreak L6-\allowbreak v2} sentence
embedding~\cite{wang2020minilm}, and reranking uses the
\texttt{cross-\allowbreak encoder/\allowbreak ms-\allowbreak marco-\allowbreak MiniLM-\allowbreak L6-\allowbreak v2}
cross-encoder. The pipeline shape
(retrieve 40 by cosine, cross-encoder rerank, keep 10) and its full
per-question token and cost ledger are reported in
Appendix~\ref{app:detail}.}:
judged tied with \inject{} on 16 of 18 document-bound questions (\inject{}
preferred on 2, \navembed{} preferred on 0), with both out-of-scope controls
also tied, at $188{,}274$ input tokens against \inject{}'s $3{,}263{,}796$,
a $17.3\times$ reduction. GTE embeddings (RUN-009) give the
largest reduction, $29.9\times$, at a lower tie rate ($15/20$; \inject{}
preferred on 3, \navembed{} on 2)~\cite{li2023gte}. Hybrid BM25 (RUN-011) is the failure mode:
length bias~\cite{robertson2009bm25} inflates the retrieved payload to
$382$K tokens (reduction falls to $8.5\times$), demonstrating that lexical
fusion can silently cost more than it returns on clause-structured text. RUN-010 (MiniLM with
ground-truth-assisted retrieval) reached $18/20$ ties but uses gold node labels
to validate retrieval, making it an oracle upper bound; it is excluded from all
cross-mode comparisons. Notably, its two \inject{} wins fall on the same two
questions as RUN-012's, consistent with the retrieval-difficulty diagnosis
below.

The two \inject{} wins in RUN-012 (\texttt{SP-first-ave-Q2} and
\texttt{SP-meta-giphy-Q3}; Appendix~\ref{app:detail}) both concern
share-purchase and disclosure-schedule provisions that draw on multiple
cross-referenced clauses. The likely cause is retrieval failure (relevant
nodes ranked outside the top-$10$ after reranking) rather than reasoning
failure, though confirming this requires the ground-truth node labels planned
in Section~\ref{sec:lessons}. This diagnosis motivated encoding
cross-references explicitly, which is exactly the signal \navindex{} navigates
(Section~\ref{sec:json}).

\textbf{Embedding model choice mattered less than descriptor richness.} Across
the Phase~II runs (Appendix~\ref{app:dev}), BGE embeddings produced more
\inject{} wins than \texttt{paraphrase-MiniLM-L6-v2} (consistent with
over-selection of lower-relevance nodes), and MPNet QA embeddings raised the
Phase~II tie count to $9/15$. But no embedding swap moved outcomes as much as
what each index node carries, the observation that motivated Comparison~2.

\textbf{Dollar cost.} \navembed{}'s recorded cost is lower than \inject{}'s in
every ledger-validated run (RUN-005 onward; Table~\ref{tab:glance}). RUN-011
is, however, the one Phase~III configuration whose per-query payload
(${\sim}19$K tokens) exceeds one-tenth of the $163$K corpus prefix, so by the
crossover rule of Section~\ref{sec:crossover} a warm-cached \inject{} would
beat it at steady state: the length-bias failure costs real money, not
only token footprint.

\FloatBarrier

\section{INJECT vs.\ NAVINDEX}
\label{sec:json}

\subsection{Motivation: descriptor richness as the binding constraint}
\label{sec:descriptor}

The \navembed{} runs identify the binding constraint: the \emph{richness} of
what each index node carries. To overcome the limitations of a vector
retrieval system, we designed \navindex{}: a retrieval system where an LLM
reasons over a compact structured index rather than searching by similarity alone.
Cosine similarity surfaces semantically similar
nodes but cannot represent structural signals such as ``this clause defines a
term,'' ``this provision contains a liability cap,'' or ``this node
cross-references Section~5.17.'' And when nodes carry only headings, the model
over-selects (R5, Section~\ref{sec:embed}). The \navindex{} format
addresses both limits by encoding each node with typed metadata and a hard
selection cap.

A vectorless, structure-navigating index of this kind is closest in spirit to
PageIndex~\cite{pageindex2025}, which builds a hierarchical table-of-contents tree
of a document and has an LLM reason over node titles and per-node summaries to
choose which sections to read. PageIndex generates a natural-language summary of
every node at index time. Our descriptor is deliberately weighted the other way:
the primary navigation signals are compact and cheap to index: headings,
clause references, and the explicit cross-reference and defined-term graph
are extracted deterministically from document structure, and the boolean
semantic flags and keyword tags are populated by Syntheia's enrichment pass
as single booleans and short tag lists rather than free text. The model
filters on these before consulting any summary text. The reason is cost.
Generating a free-text summary of every section is the expensive end of the
indexing budget and amortises only when the same document is queried many
times; a given legal instrument is often consulted only a handful of times, or
once, so that cost may never be recovered. The extracted fields carry no
generation cost at all, and the flags and tags are the cheap end of the
enrichment budget.

\subsection{Approach: the NAVINDEX format}
\label{sec:indexjson}

\textbf{Dual-file architecture.} Every processed document produces two files:
\texttt{*.index.json} (the \emph{navigation surface}: structural metadata plus a
short opening snippet per node, no full provision text) and \texttt{*.full.json}
(the \emph{provision store}: verbatim clause text fetched by node ID). The navigation
surface is designed to scale with document structure rather than text length.
The cached S1 prompt used in the scored runs serialises the six scored
documents' token-optimised indices together with the navigation instructions
and measures $98{,}562$ tokens (Appendix~\ref{app:detail}); the three large
instruments (the two LPAs and the SPA) account for roughly nine-tenths
of it, and the two amendments and the disclosure schedule for the remainder.

\textbf{Boolean pre-filter flags.} Seven semantic boolean fields (Table~\ref{tab:flags})
let the retrieval model narrow a $500$-node index to a handful of candidates in
a single pass, before reading any summary text. Config~4 also tests two
structural fields (\texttt{textLength}, \texttt{isTocEntry}) added on top of
the seven semantic flags.

\begin{table}[htb]
\centering
\caption{Boolean pre-filter flags stored in each index node.}
\label{tab:flags}
\small
\setlength{\tabcolsep}{3.5pt}
\begin{tabular}{lll}
\toprule
Field & Verbose name & Prioritise for\ldots \\
\midrule
\texttt{isDef}  & \texttt{isDefinition}      & specific defined terms \\
\texttt{hm}     & \texttt{hasMoney}          & fees, payments, thresholds \\
\texttt{hd}     & \texttt{hasDate}           & timing, terms, deadlines \\
\texttt{hpct}   & \texttt{hasPercentage}     & rates, shares, proportions \\
\texttt{hll}    & \texttt{hasLiabilityLimit} & liability caps, indemnities \\
\texttt{hc}     & \texttt{hasCondition}      & conditions precedent, triggers \\
\texttt{hpty}   & \texttt{hasParty}          & rights of a named party \\
\bottomrule
\end{tabular}
\end{table}

\textbf{Three-step navigation.}
\begin{enumerate}[label=\textbf{S\arabic*.},leftmargin=2.2em,labelsep=0.4em,itemsep=3pt,align=left]
\item \emph{Structured scan.} The LLM receives \texttt{*.index.json} as its
  cached system prompt. It applies boolean pre-filters, confirms relevance via
  titles and summaries, and returns \emph{at most ten} node IDs.
\item \emph{Deterministic fetch.} Node IDs are looked up in
  \texttt{*.full.json}; verbatim provision texts are assembled into a
  \texttt{<provisions>} block.
\item \emph{Grounded answer.} The answering model receives only the question
  and the fetched provisions (no index, no other documents).
\end{enumerate}

The hard cap of ten nodes per question bounds S3 token cost regardless of
corpus size. Prompt caching on the index block means subsequent S1 calls pay
only for the question text ($20$--$60$ tokens), not the full index, though
the model still attends over the full cached index on every scan
(Section~\ref{sec:tokens-dollars}).

\begin{table*}[!t]
\centering
\caption{\inject{} vs.\ \navindex{} across six descriptor configurations.
``Q'' is the question count for the run (Configs~1--5 predate the full
20-question set). ``Inj.~w.''/``Nav.~w.''\ are judge wins; ``Inj.\ Mtok'' and
``Nav.\ Mtok'' are total input tokens (millions); ``Red.''\ is their ratio;
dollar columns are total recorded run cost in USD. Config~6's 20 questions comprise 18
document-bound ties and 2 out-of-scope-control ties (Appendix~\ref{app:detail}).}
\label{tab:index}
\small
\setlength{\tabcolsep}{4pt}
\begin{tabular}{p{6.4cm}rrrrrrrrr}
\toprule
Configuration & Q & Ties & Inj.\ w. & Nav.\ w. & Inj.\ Mtok & Nav.\ Mtok & Red. & Nav.\ \$ & Inj.\ \$ \\
\midrule
1.\ Cross-refs $+$ summaries, all related nodes                  & 15 & 11 & 1 & 3 & 3.22 & 2.98 & 1.08$\times$ & 2.320 & 1.881 \\
2.\ Same, limited to $10$ nodes                                  & 15 & 9  & 3 & 3 & 3.22 & 2.79 & 1.15$\times$ & 1.863 & 1.881 \\
3.\ As (1), returned nodes raised to $20$                        & 15 & 10 & 3 & 2 & 3.22 & 2.84 & 1.13$\times$ & 1.913 & 1.881 \\
4.\ As (1) $+$ metadata flags\textsuperscript{$\dagger$}         & 15 & 11 & 1 & 3 & 3.22 & 3.08 & 1.05$\times$ & 2.350 & 1.881 \\
5.\ As (1), re-rank top $10$ with E5 embeddings                  & 15 & 9  & 4 & 2 & 3.22 & 2.92 & 1.10$\times$ & 2.120 & 1.881 \\
6.\ Config~1 features $+$ token-opt.\ JSON\textsuperscript{$\ddagger$}    & 20 & \textbf{20} & 0 & 0 & 3.26 & 2.03 & \textbf{1.61$\times$} & \textbf{1.213} & 1.614 \\
\bottomrule
\end{tabular}
\vspace{2pt}

\footnotesize\textsuperscript{$\dagger$}Flags: \texttt{textLength},
\texttt{isTocEntry}, \texttt{hasParty}, \texttt{hasCondition},
\texttt{hasLiabilityLimit}, \texttt{hasPercentage}, \texttt{hasDate},
\texttt{hasMoney}, \texttt{isDefinition}.\\[2pt]
\footnotesize\textsuperscript{$\ddagger$}Config~6 carries all Config~1
features (cross-references, AI-generated summaries, defined-term flags, and the
full Config~4 metadata flag set) but serialises the index with token-optimised
JSON: boolean fields are omitted when \texttt{false}, array fields when empty,
and string fields when null; key names are abbreviated to two characters; all
non-essential whitespace is stripped. These changes reduce the cached index size
by ${\sim}30\%$ without altering the navigable content, the one effect
attributable to the serialisation itself. Config~6 also moves to the full
20-question set and the Phase~III harness, so \emph{all} cross-configuration
comparisons against Configs~1--5 (ties, dollar cost, and token footprint)
carry that change;
Config~6's own headline numbers are unaffected, being paired against
\inject{} within the same run.
\end{table*}

\textbf{Cross-reference and defined-term resolution.}
Standard RAG retrieves the clause most similar to the query, but legal accuracy
requires more. A liability cap may invoke a defined term whose definition appears
20 pages earlier; an operative clause may point to a schedule that contains the
actual threshold; an amendment may silently supersede the provision the retriever
fetched. Cosine similarity and BM25 measure lexical or semantic proximity; neither
can represent these structural dependencies.

The \navindex{} format encodes them explicitly. Every node carries a
\texttt{crossReferencedIds} list naming the node IDs it references, and the
\texttt{isDefinition} flag marks defined-term nodes. The deterministic fetch
in S2 follows the cross-reference edges automatically: retrieving a clause also
retrieves every provision it cross-references, without a second similarity
search. Defined terms are handled by the flag rather than an edge: the
\texttt{isDefinition} marker lets the navigation step (S1) pull in the
definitions a question turns on, instead of leaving them to embedding
proximity. The answering model therefore receives a provision's
cross-referenced context and the relevant definitions rather than only its
highest-scoring neighbours in embedding space. This structural navigation is
not available to flat embedding retrieval; whether it is what produces
Config~6's tie result is not established (Section~\ref{sec:compare}).

\textbf{Encoding legal precedence in prompts.}
The retrieval format is only half the domain engineering; the released
prompts (Appendix~\ref{app:prompts}) carry the other half. The
answering prompts encode an amendment-precedence hierarchy (``a later-dated
amendment, side letter, restatement, or supplement \textsc{overrides} the
original to the extent of any inconsistency''), defined-term origin tracking
across documents, and an exact-string silence rule for out-of-scope questions.
The S1 navigation prompt encodes a flag-routing table (defined term
$\rightarrow$ \texttt{isDefinition}; amount $\rightarrow$ \texttt{hasMoney};
\dots) and an amendment rule: when an amendment addresses the same subject
as the original, fetch \emph{both}. These rules operationalise how
transactional lawyers actually read instruments, and they transfer to any legal
retrieval system regardless of index format.

\textbf{Descriptor ladder.}
The quality--cost trade-off maps onto a descriptor ladder of increasing richness
and indexing cost: (D1)~heading only; (D2)~$+$ structured flags;
(D3)~$+$ extracted keywords and entities; (D4)~$+$ a capped one-line
micro-summary; (D5)~full section summary. The configurations evaluated here do
not climb it one rung at a time (they already sit near the top), so
attributing answer quality to a specific rung remains an untested ablation, and
the ladder should be read as a design map rather than a result.

\subsection{Results}
\label{sec:json-results}

Table~\ref{tab:index} reports \inject{} vs.\ \navindex{} across six index
descriptor configurations. Configurations~1--5 use 15-question subsets;
Config~6 uses all 20 questions with the token-optimised index.

The configurations show three things. First, capping returned nodes at $10$
(Config~2) reduced \navindex{} cost by ${\sim}20\%$ (\$1.86 vs.\ \$2.32) with
\navindex{} wins unchanged at $3$, though ties fell from $11$ to $9$;
over-selection in Config~1 had inflated cost without protecting quality. Second,
raising the cap to $20$ (Config~3) gave no benefit: tie count dropped and
\inject{} wins rose, so the additional nodes introduce noise rather than
signal. The node cap mattered more than any descriptor enrichment we tested.
Third, the token-optimised Config~6 was judged tied with \inject{} on all 18
document-bound questions and on both out-of-scope controls, at lower dollar
cost than \inject{} (\$1.21 vs.\ \$1.61), the highest single-run tie rate in
either comparison. Because Config~6 changes the
serialisation and the question set together, comparisons against Configs~1--5
are confounded (Table~\ref{tab:index}, note~$\ddagger$); its headline numbers
are paired against \inject{} within the same run and are unaffected.

\navindex{}'s efficiency story decomposes into three numbers, all from the
per-question ledger (Appendix~\ref{app:detail}): total attended token
footprint is
$1.61\times$ lower than \inject{} (the ${\sim}98.6$K-token cached index is
re-attended on every S1 scan); the \emph{answering-model} context is
${\sim}56\times$ smaller (${\sim}2.9$K tokens of fetched provisions per query
vs.\ $163$K); and recorded dollar cost is 25\% lower (\$1.21 vs.\ \$1.61). The
long-context benefit accrues at S3, where the answer is actually composed;
the token-footprint benefit is bounded by the index size, which is why it is
$1.61\times$ rather than \navembed{}'s $17.3\times$.

\FloatBarrier
\section{Cross-Mode Discussion}
\label{sec:compare}

Table~\ref{tab:summary} places the headline numbers side by side, and
Table~\ref{tab:stats} attaches exact uncertainty to every headline rate. All
intervals are two-sided 95\% Clopper--Pearson~\cite{clopper1934} computed from
the printed counts (a convention used throughout the paper). We use this
\emph{exact} binomial method rather than the textbook normal approximation
because the counts are small ($n=15$--$20$) and several rates sit at the $0/1$
boundary, where approximate intervals become unreliable: the standard Wald
interval, for instance, collapses to zero width at a perfect $18/18$, absurdly
implying no uncertainty at all. Each interval is the range
of true rates consistent with the observed count, so a perfect $18/18$ bounds
the true tie rate only to ${\geq}0.81$, since any lower rate would make a clean
sweep implausibly lucky ($p^{18} < 0.025$).
\navembed{} was exercised across twelve configurations
(Appendix~\ref{app:dev}) and provides the
token-footprint-reduction result; \navindex{} reaches the highest single-run
tie rate
and is the only mode that navigates by structure, the signal designed for
questions that hinge on defined terms, liability caps, or explicit
cross-references, though whether that mechanism produces its tie result is
untested (Section~\ref{sec:lessons}).

A direct head-to-head comparison between RUN-012 and Config~6 is not yet
valid even though both use the full 20-question set: each mode was judged
only pairwise against its own \inject{} arm (the two modes' answers were
never judged against each other), and matching tie counts against a shared
baseline transfer token footprint and dollar cost, not quality rank. On the
present evidence the two modes occupy different points on the
quality--token-footprint
curve, and neither result supersedes the other.

\begin{table}[!htb]
\centering
\vspace{\baselineskip}
\caption{Headline configurations of the two \navigate{} modes against
\inject{} on the 20-question scored set. Each row is paired against
\inject{} only; the rows are not head-to-head comparable
(Section~\ref{sec:compare}). ``Doc-bound'' is ties\,/\,18 document-bound
questions; both out-of-scope controls were judged tied in all three runs.
``Red.''\ (reduction) is the \inject{}/\navigate{} input-token ratio.
``Ans.\ ctx.''\ is the mean answering-model context per query.}
\label{tab:summary}
\footnotesize
\setlength{\tabcolsep}{2.6pt}
\begin{tabular}{p{2.55cm}ccccc}
\toprule
Mode & Doc-bound & Red. & Ans.\ ctx. & Nav.\ \$ & Inj.\ \$ \\
\midrule
\navembed{} \mbox{RUN-012} (sem.$+$rerank) & 16/18 & $17.3\times$ & 9.4K & 0.64 & 1.61 \\
\navembed{} \mbox{RUN-009} (GTE)          & 13/18 & $29.9\times$ & 5.5K & 0.41 & 1.62 \\
\navindex{}\newline \mbox{Config-6}\newline (token-opt.)       & \textbf{18/18} & $1.61\times$ & \textbf{2.9K} & 1.21 & 1.61 \\
\bottomrule
\end{tabular}
\end{table}

\begin{table}[!htb]
\centering
\caption{Exact uncertainty for the headline rates (Clopper--Pearson intervals,
as described in Section~\ref{sec:compare}). ``Forced-decision'' is the
ties-disallowed probe of Section~\ref{sec:threats}: the share of forced
verdicts awarded to \inject{}.}
\label{tab:stats}
\footnotesize
\setlength{\tabcolsep}{3.2pt}
\begin{tabular*}{\linewidth}{@{\extracolsep{\fill}}lcc}
\toprule
Quantity & Count & 95\% CI \\
\midrule
RUN-012 ties, document-bound        & 16/18 & $[0.65, 0.99]$ \\
RUN-012 ties, full set              & 18/20 & $[0.68, 0.99]$ \\
RUN-012 \inject{} wins              & 2/18  & $[0.01, 0.35]$ \\
Config~6 ties, document-bound       & 18/18 & $[0.81, 1.00]$ \\
Config~6 ties, full set             & 20/20 & $[0.83, 1.00]$ \\
RUN-009 ties, document-bound        & 13/18 & $[0.47, 0.90]$ \\
RUN-009 ties, full set              & 15/20 & $[0.51, 0.91]$ \\
Forced-decision \inject{} share     & 10/15 & $[0.38, 0.88]$ \\
Judge audit ``likely wrong''        & 2/15  & $[0.02, 0.40]$ \\
\bottomrule
\end{tabular*}
\end{table}

\twocolumn[{%
\section{Cost Analysis: The Caching Crossover}
\label{sec:crossover}
\noindent
\includegraphics[width=0.49\textwidth]{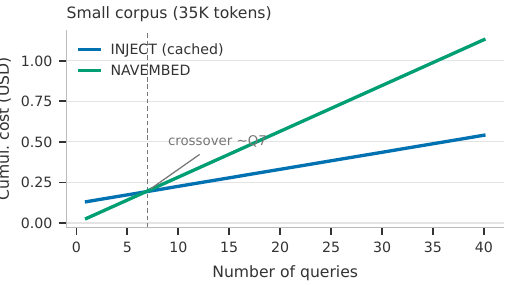}\hfill
\includegraphics[width=0.49\textwidth]{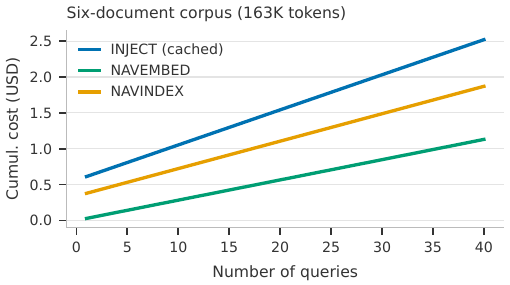}
\setlength{\abovecaptionskip}{3pt}%
\captionof{figure}{Cumulative dollar cost versus repeated queries; \navembed{}
at the measured RUN-012 payload ($9.4$K tokens/query, full input price).
\textbf{Left:} a single ${\sim}35$K-token document (\navindex{} omitted; its
cost depends on that document's index size). \textbf{Right:} the measured
$163$K six-document corpus; \navindex{} at its Config~6 structure (cached
$98.6$K index re-read per query plus ${\sim}2.9$K provisions at full input
price; Eq.~\ref{eq:crossover-json}).}
\label{fig:crossover}
\vspace{0.3em}
}]

Phase~III settled the token-footprint comparison; the dollar comparison is
less direct.
The price model
uses base input at \$3.00 per million tokens ($p_{\text{in}}$), cache write at
\$3.75 ($p_{\text{w}}$), cache read at \$0.30 ($p_{\text{r}}$), and output at
\$15~\cite{anthropic2026pricing}. Output tokens are excluded from the cost model below because
output length did not differ systematically between \inject{} and \navigate{}
modes in these runs (median output length within $\pm 8\%$ across paired
answers); if output length diverges substantially in other corpora or question
sets, the output term should be reinstated. Every aggregate token and dollar
figure for the two headline runs reconciles to the per-question ledgers in
Appendix~\ref{app:detail} under this price card; dollar accounting is
ledger-validated from RUN-005 onward (Phase~I dollar cells were recorded under
an earlier harness and are excluded; Appendix~\ref{app:dev}).
For a corpus of $C$ tokens queried repeatedly against a \navigate{}
payload of $R$ tokens, at warm steady state:
\begin{equation}
c_{\inject{}}^{\text{warm}} = C\,p_{\text{r}},
\qquad
c_{\navigate{}} = R\,p_{\text{in}} .
\end{equation}
\inject{} is cheaper per query exactly when
\begin{equation}
C < R\,\frac{p_{\text{in}}}{p_{\text{r}}}
= 10\,R .
\label{eq:crossover}
\end{equation}
\emph{Practical rule for \textup{\navembed{}}:} cached \inject{} is cheaper
in dollars only when
the corpus stays below roughly ten times the retrieval payload. The headline
RUN-012 payload averages ${\sim}9.4$K tokens per query, putting break-even
near $94$K tokens; across the Phase~III configurations (payloads
$5.5$K--$19.1$K per query) the threshold spans roughly $55$K--$191$K. The
measured six-document prefix of $163{,}147$ tokens exceeds the RUN-012
threshold, so \navembed{} stays cheaper at steady state, except for the
BM25 configuration (RUN-011), whose inflated $19.1$K payload pushes its
threshold past the corpus size (Section~\ref{sec:embed-results}).

\navindex{} has a different cost profile
(steps as defined in Section~\ref{sec:indexjson}): its S1 index
(${\sim}98.6$K tokens) is itself cached, so only the small S3
provision payload (${\sim}2.9$K tokens per query) is billed at full input
price. Warm-state per-query cost is $I\,p_{\text{r}} + S\,p_{\text{in}}$, where
$I$ is the cached index size and $S$ the provision payload. The crossover
condition becomes
\begin{multline}
C\,p_{\text{r}} < I\,p_{\text{r}} + S\,p_{\text{in}}
\;\Longrightarrow\;
C < I + S\,\frac{p_{\text{in}}}{p_{\text{r}}} \\[2pt]
\approx 98{,}600 + 2{,}900\times 10
\approx 128{,}000 \text{ tokens.}
\label{eq:crossover-json}
\end{multline}
Cached \inject{} beats \navindex{} only when the corpus stays below roughly
$128$K tokens. Both thresholds are exceeded by the measured $163$K
six-document prefix, matching the recorded outcome (\$1.21 vs.\ \$1.61); for a
single ${\sim}35$K-token document, \inject{} is the cheaper steady-state
option against either \navigate{} mode. (The Phase~II corpus
was a different six-document sample of the workspace (it carried the
PPG/Comex SPA in place of the disclosure schedule) at ${\sim}214.5$K
tokens per question against the scored corpus's $163.1$K.)

\textbf{Session economics.} The warm-state equation omits the one-time cache
write, which must also be amortised, and Anthropic's prompt cache has a
five-minute TTL~\cite{anthropic2024caching}. Queries arriving more than five
minutes apart invalidate the cache entry, and the next call pays the write
premium again. In a synchronous review session (many questions in one sitting)
the model approximates reality; in an asynchronous workflow (a lawyer asking
three questions today and two tomorrow) the cache write may be paid for every
session, shifting the crossover to a higher query count or removing it
entirely for infrequent use. Cached \inject{}'s cheapness below the threshold
therefore assumes a usage pattern that legal review often violates. The
analysis here assumes a single warm-cache session.
Figure~\ref{fig:crossover} shows cumulative costs for two corpus sizes: for a
single ${\sim}35$K-token document (left), cached \inject{} overtakes
\navembed{} after about seven repeated questions; for the $163$K
six-document corpus (right), \navembed{} remains cheaper at every repetition
count.

\section{Threats to Validity}
\label{sec:threats}

\textbf{What a tie certifies.} A tie under the reference-anchored protocol
(Section~\ref{sec:judging}) certifies that the judge found both answers
equally consistent with the verified ground truth, not that either answer
is verified correct. Judge error is bounded by the audit ($2/15$ verdicts
likely wrong; Table~\ref{tab:stats}). A forced-decision variant of the
\navindex{} Config~2 setup (ties disallowed) on the 15-question Phase~II
pool, whose DocNavBench questions are disjoint from the scored set
(Section~\ref{sec:method}), showed a $10$--$5$ \inject{} preference,
judged with the Phase~I--II preference judge. That lean is statistically
indistinguishable from chance (two-sided exact binomial test~\cite{clopper1934}, $p = 0.30$), the
interval ($[0.38, 0.88]$) is wide, and it attaches to the Phase~II
questions rather than the scored set. Even so, it remains the only
forced-decision datum available.
Single-pass tie counts also carry judge-pass noise: two stored judging
passes over the \emph{identical} RUN-010 answers differed on one verdict in
twenty (a control-question tie versus a \navigate{} win), so tallies should
be read with at least $\pm 1$ verdict of judging noise, well inside the
intervals of Table~\ref{tab:stats}. The forced-decision probe has not been
re-run under the final protocol or on the full 20 questions; a
forced-decision run on Config~6 is the single most important follow-up, since
it would establish whether the headline tie counts reflect genuine
interchangeability or differences beneath the judge's resolution.

\textbf{Judge--answerer provider overlap.} The judge
(\texttt{claude-opus-4-7}) shares a provider with the answering model
(\texttt{claude-sonnet-4-6}). Because \emph{both} arms of every pair are
answered by the same model, any family-level self-preference~\cite{panickssery2024self} applies to both
answers symmetrically: the comparison is between retrieval conditions, not
model families, so provider bias cannot systematically favour one arm. The
residual risk is tie \emph{inflation} through stylistic coherence,
compounded by the reference answers being model-drafted before human
verification (Section~\ref{sec:method}), which the reference-anchored
judging criteria (verdicts keyed to key facts against the reference, with
explicit instructions to ignore formatting, length, and tone) and the
forced-decision probe partially address, and which the audit bounds
empirically. A cross-provider judge remains future work.

\textbf{No ground-truth node labels.} Retrieval recall and precision cannot be
measured. When \navigate{} answers ``not found,'' it is impossible to tell
whether it fetched the wrong nodes or the right nodes that lacked the answer;
the diagnosis of RUN-012's two \inject{} wins
(Section~\ref{sec:embed-results}) is correspondingly conjectural.

\textbf{Corpus-selection artefacts.} In one Phase~II run, $11/15$ questions
were unanswerable because amendments were sampled without their base
documents. Both methods return ``not in documents'' and are scored a tie,
a phantom tie that inflates apparent convergence. The artefact predates the
scored set: the printed 15-question rows are inconsistent with it
(their per-question records show substantive retrieval and decisive verdicts
on the same questions), and the scored runs are protected by
construction rather than by luck. Each document-bound question has a verified
ground-truth answer authored against the indexed documents, and the
per-question ledgers (Appendix~\ref{app:detail}) show substantive answers on
all 18 with no mass ``not found'' behaviour; only the 2 out-of-scope controls
are deliberately unanswerable.

\textbf{Out-of-scope controls.} The stored answers verify declination
directly for the shared \inject{} arm (reused across RUN-010--012): both control answers are
the mandated refusal string. The stored \navigate{} answers in the
ground-truth-judged oracle run likewise open with the exact refusal; for
RUN-012's and Config~6's \navigate{} arms the controls are certified as
\emph{judged tied} against a reference that says the answer is absent, under
a criterion where a fabricated answer loses.

\textbf{Difficulty profile and phrasing.} The internal difficulty labels
(14 high, 6 medium; Appendix~\ref{app:questions}) rate retrieval dispersion
(how widely the answer spreads across clauses and documents), not
cognitive difficulty: the reviewing lawyer assessed only a handful of the
20 as genuinely hard, the majority being precise
retrieval-and-comprehension questions a modern LLM answers reliably once the
governing text is in context. That cuts both ways. It makes the comparison a clean test of the retrieval
condition, since ties are the expected outcome when both arms receive the
right provisions and the informative signal is the failures; but the
evaluation says nothing about legal reasoning. The
lawyer also noted the questions are phrased answer-aware, written as if the
asker already knows what the document contains, unlike DocNavBench's
issue-driven register (Section~\ref{sec:method}), so the phrasing may
flatter retrieval relative to real-world queries.

\textbf{Scale.} All results come from one answering model, one judge, and one
run per configuration, on a 20-question set over six documents;
Table~\ref{tab:stats} quantifies the resulting uncertainty, and the wide
intervals are the honest reading of every rate in this paper. The \navembed{}
and \navindex{} comparisons are asymmetric in configuration count (twelve
vs.\ six), and only their strongest configurations share the scored question
set, so the cross-mode reading (Section~\ref{sec:compare}) is indicative, not
definitive.

\section{Lessons Learned and Future Work}
\label{sec:lessons}

Four operational lessons transfer beyond this system. \textbf{(1) Cap the
selection.} The ten-node cap cut \navindex{} cost by ${\sim}20\%$ with wins
unchanged (ties dipped $11{\to}9$; Table~\ref{tab:index}), and every uncapped
or weakly-signalled variant (titles-only scan, 20-node cap, all-related-nodes)
over-selected; a hard cap is the single highest-leverage retrieval control we
found. \textbf{(2) Lexical fusion can
silently cost money.} BM25 length bias inflated the hybrid payload
$2\times$ over pure semantic retrieval and pushed the configuration past the
caching crossover, worse than \inject{} at steady state. \textbf{(3) What a
node carries matters more than which embedding scores it.} Embedding swaps
moved outcomes less than descriptor enrichment across Phases~I--II
(Appendix~\ref{app:dev}), which is what motivated the \navindex{} format.
\textbf{(4)
Track token footprint and dollar cost separately.} Prompt caching makes
dollar comparisons corpus- and session-dependent
(Section~\ref{sec:crossover}); token-footprint reduction is the durable
quantity.

The evaluation gaps, in order of urgency: a forced-decision judging run on
Config~6 (Section~\ref{sec:threats}); ground-truth node labels to measure
retrieval recall directly and to diagnose the two RUN-012 losses; a
head-to-head judging run in which the two \navigate{} modes' answers are
compared directly rather than each against \inject{};
evaluation on the 52-question open-search DocNavBench set; substring rubrics
(\texttt{must\_contain}) as a judge-free scoring dimension; easy single-fact
questions to establish a retrieval floor; corpus pairing that always includes
base agreements with amendments; questions that require reconciling an
original agreement against its amendments across documents (a cross-instrument
dependency the present question set does not isolate); a second-annotator pass over the headline
runs' judge verdicts; a cross-provider judge; and integrating a
LEDGAR-style clause-type classifier~\cite{tuggener2020ledgar} as an index
field.

\section{Conclusion}
\label{sec:conclusion}

\inject{} (placing the entire corpus in context) requires no retriever,
and on a small corpus queried repeatedly within a warm cache session it is the
cheaper option in dollars by the crossover rule of
Section~\ref{sec:crossover}. Outside that regime, both structured
alternatives held answer quality at a fraction of the token footprint on this
evaluation, with the caveats of
Section~\ref{sec:threats}.

Against \navembed{}, semantic retrieval with reranking (RUN-012) was judged
tied with \inject{} on 16 of 18 document-bound questions
(\inject{} preferred on 2) while attending to $17.3\times$ fewer input tokens
($188$K vs.\ $3.26$M); GTE retrieval pushed the reduction to $29.9\times$ at a
lower tie rate ($15/20$). Across all twelve configurations the
reduction never fell below $8.5\times$, and the binding constraint on quality
was descriptor richness, not embedding choice.

Against \navindex{}, its index format encodes the defined-term and
cross-reference relationships a legal answer depends on as traversable edges,
and the deterministic fetch follows them without a second similarity search.
Its token-optimised configuration was judged tied with \inject{}
on all 18 document-bound questions (the highest single-run tie rate in
either comparison) at $1.61\times$ lower total token footprint, a
${\sim}56\times$ smaller answering context, and 25\% lower recorded dollar
cost
(\$1.21 vs.\ \$1.61).

For practitioners the decision rule is short: below ${\sim}10\times$ the
retrieval payload (a constant set by the cache-read discount
$p_{\text{in}}/p_{\text{r}}$ under the Section~\ref{sec:crossover} price
card), inject and cache; above it, retrieve, with \navembed{} when maximal
token-footprint reduction matters and \navindex{} when structural fidelity
does. Real transaction sets outgrow the context window, where
\inject{} stops being an option at all, and asynchronous legal workflows
re-pay the cache write per session, eroding cached \inject{}'s advantage even
below the threshold. Structured retrieval is how the system holds quality
when the corpus keeps growing.

\subsection*{Acknowledgements}

We are grateful to Horace Wu, who authored the 52-question DocNavBench
open-search set and verified the scored benchmark's reference answers against
the source documents, and to \mbox{Mahmoud Abdallah} for his work on the
document-index API and surrounding features.

\bibliographystyle{unsrt}

\begin{thebibliography}{99}
\setlength{\itemsep}{2pt}

\bibitem{anthropic2024mcp}
Anthropic.
\newblock Introducing the Model Context Protocol.
\newblock Anthropic, November 2024.
\newblock \url{https://www.anthropic.com/news/model-context-protocol}

\bibitem{liu2024lost}
N.~F.~Liu, K.~Lin, J.~Hewitt, A.~Paranjape, M.~Bevilacqua, F.~Petroni, P.~Liang.
\newblock Lost in the Middle: How Language Models Use Long Contexts.
\newblock \emph{TACL}, 2024.

\bibitem{hendrycks2021cuad}
D.~Hendrycks, C.~Burns, A.~Chen, S.~Ball.
\newblock CUAD: An Expert-Annotated NLP Dataset for Legal Contract Review.
\newblock In \emph{NeurIPS Datasets and Benchmarks}, 2021.

\bibitem{koreeda2021contractnli}
Y.~Koreeda, C.~D.~Manning.
\newblock ContractNLI: A Dataset for Document-Level Natural Language Inference
for Contracts.
\newblock In \emph{Findings of EMNLP}, 2021.

\bibitem{merola2025chunking}
C.~Merola, J.~Singh.
\newblock Reconstructing Context: Evaluating Advanced Chunking Strategies for
Retrieval-Augmented Generation.
\newblock \emph{arXiv:2504.19754}, 2025.

\bibitem{taiwo2026chunking}
S.~Taiwo, M.~A.~Yusoff.
\newblock Evaluating Chunking Strategies for Retrieval-Augmented Generation in
Oil and Gas Enterprise Documents.
\newblock \emph{arXiv:2603.24556}, 2026.

\bibitem{lewis2020rag}
P.~Lewis, E.~Perez, A.~Piktus, F.~Petroni, V.~Karpukhin, N.~Goyal, et al.
\newblock Retrieval-Augmented Generation for Knowledge-Intensive NLP Tasks.
\newblock In \emph{NeurIPS}, 2020.

\bibitem{gao2023survey}
Y.~Gao, Y.~Xiong, X.~Gao, K.~Jia, J.~Pan, Y.~Bi, et al.
\newblock Retrieval-Augmented Generation for Large Language Models: A Survey.
\newblock \emph{arXiv:2312.10997}, 2023.

\bibitem{karpukhin2020dpr}
V.~Karpukhin, B.~O\u{g}uz, S.~Min, P.~Lewis, L.~Wu, S.~Edunov, D.~Chen, W.~Yih.
\newblock Dense Passage Retrieval for Open-Domain Question Answering.
\newblock In \emph{EMNLP}, 2020.

\bibitem{reimers2019sbert}
N.~Reimers, I.~Gurevych.
\newblock Sentence-BERT: Sentence Embeddings using Siamese BERT-Networks.
\newblock In \emph{EMNLP}, 2019.

\bibitem{wang2022e5}
L.~Wang, N.~Yang, X.~Huang, B.~Jiao, L.~Yang, D.~Jiang, R.~Majumder, F.~Wei.
\newblock Text Embeddings by Weakly-Supervised Contrastive Pre-training.
\newblock \emph{arXiv:2212.03533}, 2022.

\bibitem{li2023gte}
Z.~Li, X.~Zhang, Y.~Zhang, D.~Long, P.~Xie, M.~Zhang.
\newblock Towards General Text Embeddings with Multi-stage Contrastive Learning.
\newblock \emph{arXiv:2308.03281}, 2023.

\bibitem{xiao2023bge}
S.~Xiao, Z.~Liu, P.~Zhang, N.~Muennighoff, D.~Lian, J.-Y.~Nie.
\newblock C-Pack: Packaged Resources to Advance General Chinese Embedding.
\newblock \emph{arXiv:2309.07597}, 2023.

\bibitem{song2020mpnet}
K.~Song, X.~Tan, T.~Qin, J.~Lu, T.-Y.~Liu.
\newblock MPNet: Masked and Permuted Pre-training for Language Understanding.
\newblock In \emph{NeurIPS}, 2020.

\bibitem{robertson2009bm25}
S.~Robertson, H.~Zaragoza.
\newblock The Probabilistic Relevance Framework: BM25 and Beyond.
\newblock \emph{Foundations and Trends in Information Retrieval}, 3(4), 2009.

\bibitem{cormack2009rrf}
G.~V.~Cormack, C.~L.~A.~Clarke, S.~Buettcher.
\newblock Reciprocal Rank Fusion Outperforms Condorcet and Individual Rank
Learning Methods.
\newblock In \emph{SIGIR}, 2009.

\bibitem{nogueira2019rerank}
R.~Nogueira, K.~Cho.
\newblock Passage Re-ranking with BERT.
\newblock \emph{arXiv:1901.04085}, 2019.

\bibitem{anthropic2024caching}
Anthropic.
\newblock Prompt Caching with Claude.
\newblock Anthropic Documentation, 2024.
\newblock \url{https://docs.anthropic.com/en/docs/build-with-claude/prompt-caching} [Accessed: June 2026; content subject to change as a living document].

\bibitem{zheng2023judge}
L.~Zheng, W.-L.~Chiang, Y.~Sheng, S.~Zhuang, Z.~Wu, Y.~Zhuang, et al.
\newblock Judging LLM-as-a-Judge with MT-Bench and Chatbot Arena.
\newblock In \emph{NeurIPS}, 2023.

\bibitem{wang2023fair}
P.~Wang, L.~Li, L.~Chen, Z.~Cai, D.~Zhu, B.~Lin, Y.~Cao, et al.
\newblock Large Language Models are not Fair Evaluators.
\newblock \emph{arXiv:2305.17926}, 2023.

\bibitem{tuggener2020ledgar}
D.~Tuggener, P.~von~D\"aniken, T.~Peetz, M.~Cieliebak.
\newblock LEDGAR: A Large-Scale Multi-label Corpus for Text Classification of
Legal Provisions in Contracts.
\newblock In \emph{LREC}, 2020.

\bibitem{sarthi2024raptor}
P.~Sarthi, S.~Abdullah, A.~Tuli, S.~Khanna, A.~Goldie, C.~D.~Manning.
\newblock RAPTOR: Recursive Abstractive Processing for Tree-Organized Retrieval.
\newblock In \emph{ICLR}, 2024.

\bibitem{pageindex2025}
M.~Zhang, Y.~Tang, et al.\ (VectifyAI).
\newblock PageIndex: Document Index for Vectorless, Reasoning-based RAG.
\newblock 2025.
\newblock \url{https://github.com/VectifyAI/PageIndex}

\bibitem{clopper1934}
C.~J.~Clopper, E.~S.~Pearson.
\newblock The Use of Confidence or Fiducial Limits Illustrated in the Case of
the Binomial.
\newblock \emph{Biometrika}, 26(4):404--413, 1934.

\bibitem{wang2020minilm}
W.~Wang, F.~Wei, L.~Dong, H.~Bao, N.~Yang, M.~Zhou.
\newblock MiniLM: Deep Self-Attention Distillation for Task-Agnostic Compression
of Pre-Trained Transformers.
\newblock In \emph{NeurIPS}, 2020.

\bibitem{anthropic2026pricing}
Anthropic.
\newblock Pricing.
\newblock Anthropic, 2026.
\newblock \url{https://www.anthropic.com/pricing} [Accessed: June 2026; prices subject to change].

\bibitem{panickssery2024self}
A.~Panickssery, S.~R.~Bowman, S.~Feng.
\newblock LLM Evaluators Recognize and Favor Their Own Generations.
\newblock In \emph{NeurIPS}, 2024.

\end{thebibliography}

\balance

\onecolumn
\appendix
\raggedbottom  
\section{Run History and Lessons}
\label{app:dev}

The twelve \navembed{} configurations were run in three phases as
the evaluation harness matured. Each run is an internally controlled paired
comparison (\inject{} vs.\ \navembed{} on identical questions), but five
things changed \emph{between} phases: the question pool (the DocNavBench
subset in Phases~I--II vs.\ the scored set in Phase~III;
Section~\ref{sec:method}), the corpus sample (Phase~II carried the PPG/Comex
SPA in place of the disclosure schedule; Section~\ref{sec:crossover}), the
prompts, the caching setup (integrated at RUN-005), and the judge (the
Phase~I--II preference judge, Appendix~\ref{app:p-judge-pref}, vs.\ the reference-anchored
judge, Appendix~\ref{app:p-judge-gt}; Section~\ref{sec:judging}). Cross-run tie counts are
therefore run history rather than a controlled sweep; the paper's
claims rest on the Phase~III runs (Section~\ref{sec:embed-results}). Footprint reduction (the deterministic
quantity) held at $8.5$--$29.9\times$ across all twelve runs.

\vspace{\fill}
\begin{table}[H]
\centering
\caption{All scored-run \navembed{} configurations. ``Red.''\ (reduction) is the
\inject{}/\navembed{} input-token ratio. ``Inj.~w.''\ and ``Nav.~w.''\ are
judge wins for \inject{} and \navembed{}. Costs are recorded operational spend
in USD; Phase~I cells (RUN-001--004) are omitted because they were recorded
under an earlier harness and do not reconcile with the
Section~\ref{sec:crossover} price model (dollar accounting is ledger-validated
from RUN-005 onward; Appendix~\ref{app:detail}).
Win splits are taken from the per-question run records (the run-time
ledgers; RUN-010 from its stored run file, in which both judge directions
agree on every question).}
\label{tab:glance}
\small
\setlength{\tabcolsep}{4pt}
\begin{tabular*}{\linewidth}{@{\extracolsep{\fill}}llrrrrrrrrrr}
\toprule
Run & Method & Q & Ties & Inj.~w. & Nav.~w. & Inj.\ tokens & Nav.\ tokens & Red. & Inj.\ \$ & Nav.\ \$ \\
\midrule
RUN-001 & Embed $+$ filter prompt      & 3  & 3  & 0 & 0 & 347{,}107     & 22{,}957      & 15.1$\times$ & -- & -- \\
RUN-002 & Embed, no filter reasoning   & 3  & 3  & 0 & 0 & 347{,}107     & 23{,}086      & 15.0$\times$ & -- & -- \\
RUN-003 & Embed $+$ cross-refs         & 3  & 2  & 0 & 1 & 347{,}107     & 23{,}625      & 14.7$\times$ & -- & -- \\
RUN-004 & Embed, 9 questions           & 9  & 6  & 0 & 3 & 1{,}041{,}298 & 104{,}975     & 9.9$\times$  & -- & -- \\
RUN-005 & Caching integrated           & 15 & 7  & 6 & 2 & 3{,}217{,}792 & 171{,}307     & 18.8$\times$ & 1.881 & 0.611 \\
RUN-006 & Prompt revision              & 15 & 7  & 7 & 1 & 3{,}217{,}792 & 173{,}527     & 18.5$\times$ & 1.881 & 0.632 \\
RUN-007 & Top-$k$ embed $+$ cross-ref  & 15 & 5  & 8 & 2 & 3{,}217{,}792 & 231{,}486     & 13.9$\times$ & 1.881 & 0.818 \\
RUN-008 & QA embeddings (MPNet)        & 15 & 9  & 5 & 1 & 3{,}217{,}792 & 203{,}295     & 15.8$\times$ & 1.881 & 0.744 \\
RUN-009 & Scored Q-set, GTE embed      & 20 & 15 & 3 & 2 & 3{,}263{,}796 & 109{,}214     & \textbf{29.9$\times$} & 1.621 & 0.406 \\
RUN-010\textsuperscript{$\ddagger$} & MiniLM $+$ GT answers & 20 & 18 & 2 & 0 & 3{,}263{,}796 & 174{,}310 & 18.7$\times$ & 1.614 & 0.601 \\
RUN-011 & Hybrid semantic $+$ BM25     & 20 & 16 & 4 & 0 & 3{,}263{,}796 & 382{,}240     & 8.5$\times$  & 1.614 & 1.217 \\
RUN-012 & Semantic $+$ rerank          & 20 & \textbf{18} & \textbf{2} & \textbf{0} & 3{,}263{,}796 & 188{,}274 & 17.3$\times$ & 1.614 & 0.644 \\
\bottomrule
\end{tabular*}
\vspace{2pt}

\footnotesize\textsuperscript{$\ddagger$}Oracle configuration: uses
ground-truth node labels to validate retrieval at indexing time; excluded from
cross-mode comparison (\S\ref{sec:compare}).
\end{table}

\vspace{\fill}
\textbf{Phase~I (RUN-001--004): harness construction.} The first runs
built and validated the harness on 3--9 questions from the Phase~I--II pool
(a 15-question DocNavBench subset; Section~\ref{sec:method}):
filtering irrelevant nodes, stripping rationale from model output, and adding
cross-reference expansion. \navembed{} already used $10$--$15\times$ fewer
tokens than \inject{}; RUN-003 produced the first outright \navembed{} win,
and RUN-004 closed the phase with three \navembed{} wins and six ties.

\vspace{\fill}
\textbf{Phase~II (RUN-005--008): caching and embeddings.} Moving to the
full 15-question pool with prompt caching integrated, the token
footprint
held $14$--$19\times$ below \inject{}; embedding swaps were tried here
(Section~\ref{sec:embed-results}): \texttt{paraphrase-MiniLM-L6-v2} (the
incumbent default), \texttt{intfloat/e5-base-v2}, and
\texttt{BAAI/bge-base-en-v1.5} were compared on the same pool, and MPNet QA
embeddings (\texttt{multi-qa-mpnet-base-cos-v1}, RUN-008) gave the best
Phase~II tie count ($9/15$).

\vspace{\fill}
\textbf{Phase~III (RUN-009--012): the scored comparison.} On the full
20-question set with the final harness, outcomes are reported in
Section~\ref{sec:embed-results}. The \inject{} wins concentrated in
Phase~II, while the harness was still being tuned (Table~\ref{tab:glance},
Inj.~w.\ column); this trend is an engineering observation, not a
measured result, since the question sets differ across phases.

\vspace{\fill}
\clearpage
\section{Per-Question Detail for Two Representative Runs}
\label{app:detail}

Tables~\ref{tab:cfg6} and~\ref{tab:exp012} give per-question token and cost breakdowns.
Column key: \textbf{out}~= output tokens; \textbf{CW}~= cache-write (Q1 only); \textbf{CR}~= cache-read (Q2--Q20); \textbf{in}~= non-cached input; \textbf{Inj}/\textbf{Nav}~= cost in USD\@.
Every aggregate token and dollar figure reported for these two runs in the body
reconciles to these ledgers under the Section~\ref{sec:crossover} price card.
Two bookkeeping notes: the \inject{} token totals in Table~\ref{tab:glance} and
Table~\ref{tab:index} additionally include $856$ non-cached question-text input
tokens not broken out as a column here; and the Nav rows likewise include $20$--$60$
tokens of non-cached S1 question-text input per query not broken out as a
column.

\begin{table}[H]
\centering
\caption{Config~6: \navindex{} token-optimised, 20 questions (all ties).
  Nav.\ S1 caches the $\sim$98.6\,K-token index JSON (CW on Q1, CR on Q2--Q20);
  Nav.\ S3 fetches verbatim clauses (no caching).}
\label{tab:cfg6}
\renewcommand{\arraystretch}{1.05}
\footnotesize
\setlength{\tabcolsep}{2.5pt}
\begin{tabular*}{\linewidth}{@{\extracolsep{\fill}}lrrrrrrrrrrl}
\toprule
& \multicolumn{3}{c}{Inject} & \multicolumn{3}{c}{Nav.\ S1} & \multicolumn{2}{c}{Nav.\ S3} & \multicolumn{2}{c}{Cost (\$)} & \\
\cmidrule(lr){2-4}\cmidrule(lr){5-7}\cmidrule(lr){8-9}\cmidrule(lr){10-11}
QID & out & CW & CR & out & CW & CR & in & out & Inj & Nav & V \\
\midrule
FA-fluor-am1-Q1  & 170 & 163{,}147 &         0 &  50 & 98{,}562 &         0 &  1{,}621 & 149 & 0.6144 & 0.3776 & tie \\
FA-fluor-am1-Q2  & 218 &         0 & 163{,}147 & 239 &        0 & 98{,}562 &  4{,}115 & 135 & 0.0523 & 0.0476 & tie \\
FA-fluor-am1-Q3  & 123 &         0 & 163{,}147 & 113 &        0 & 98{,}562 &  2{,}085 & 269 & 0.0509 & 0.0417 & tie \\
FA-generac-Q1    & 129 &         0 & 163{,}147 &  28 &        0 & 98{,}562 &  1{,}057 & 164 & 0.0510 & 0.0357 & tie \\
FA-generac-Q2    &  94 &         0 & 163{,}147 &  28 &        0 & 98{,}562 &    990 &  47 & 0.0504 & 0.0338 & tie \\
FA-generac-Q3    & 208 &         0 & 163{,}147 &  88 &        0 & 98{,}562 &  1{,}377 & 284 & 0.0522 & 0.0394 & tie \\
LP-thomas-Q1     & 304 &         0 & 163{,}147 & 197 &        0 & 98{,}562 &  8{,}547 & 327 & 0.0536 & 0.0632 & tie \\
LP-thomas-Q2     & 354 &         0 & 163{,}147 &  71 &        0 & 98{,}562 &  2{,}317 & 422 & 0.0544 & 0.0441 & tie \\
LP-thomas-Q3     & 385 &         0 & 163{,}147 & 197 &        0 & 98{,}562 &  3{,}636 & 631 & 0.0549 & 0.0531 & tie \\
LP-carlyle-Q1    & 326 &         0 & 163{,}147 & 118 &        0 & 98{,}562 & 12{,}840 & 507 & 0.0540 & 0.0776 & tie \\
LP-carlyle-Q2    & 242 &         0 & 163{,}147 &  30 &        0 & 98{,}562 &  1{,}813 & 182 & 0.0527 & 0.0383 & tie \\
LP-carlyle-Q3    & 331 &         0 & 163{,}147 &  30 &        0 & 98{,}562 &  1{,}176 & 161 & 0.0540 & 0.0360 & tie \\
SP-first-ave-Q1  & 296 &         0 & 163{,}147 &  52 &        0 & 98{,}562 &  1{,}114 & 247 & 0.0535 & 0.0375 & tie \\
SP-first-ave-Q2  & 147 &         0 & 163{,}147 &  30 &        0 & 98{,}562 &  1{,}299 & 146 & 0.0513 & 0.0363 & tie \\
SP-first-ave-Q3  & 218 &         0 & 163{,}147 &  30 &        0 & 98{,}562 &  1{,}219 & 166 & 0.0524 & 0.0363 & tie \\
SP-meta-giphy-Q1 & 372 &         0 & 163{,}147 & 169 &        0 & 98{,}562 &  1{,}842 & 535 & 0.0546 & 0.0458 & tie \\
SP-meta-giphy-Q2 & 170 &         0 & 163{,}147 & 192 &        0 & 98{,}562 &  5{,}699 & 198 & 0.0516 & 0.0527 & tie \\
SP-meta-giphy-Q3 & 340 &         0 & 163{,}147 &  77 &        0 & 98{,}562 &  1{,}266 & 338 & 0.0541 & 0.0397 & tie \\
out-of-scope-Q1  &  12 &         0 & 163{,}147 &  29 &        0 & 98{,}562 &  2{,}086 &  71 & 0.0492 & 0.0374 & tie \\
out-of-scope-Q2  & 206 &         0 & 163{,}147 & 136 &        0 & 98{,}562 &  1{,}782 & 160 & 0.0522 & 0.0395 & tie \\
\midrule
Total & 4{,}645 & 163{,}147 & 3{,}099{,}793 & 1{,}904 & 98{,}562 & 1{,}872{,}678 & 57{,}881 & 5{,}139 & 1.6140 & 1.2133 & 20t \\
\bottomrule
\end{tabular*}

\vspace{0.6em}

\caption{RUN-012: semantic $+$ rerank, 20 questions (18 ties, 2 inject wins).
  Navigate retrieves top-$k$ chunks via embedding $+$ rerank; no caching.}
\label{tab:exp012}
\renewcommand{\arraystretch}{1.05}
\footnotesize
\setlength{\tabcolsep}{2.5pt}
\begin{tabular*}{\linewidth}{@{\extracolsep{\fill}}lrrrrrrrl}
\toprule
& \multicolumn{3}{c}{Inject} & \multicolumn{2}{c}{Navigate} & \multicolumn{2}{c}{Cost (\$)} & \\
\cmidrule(lr){2-4}\cmidrule(lr){5-6}\cmidrule(lr){7-8}
QID & out & CW & CR & in & out & Inj & Nav & V \\
\midrule
FA-fluor-am1-Q1  & 170 & 163{,}147 &         0 & 17{,}483 & 145 & 0.6144 & 0.0546 & tie \\
FA-fluor-am1-Q2  & 218 &         0 & 163{,}147 &  5{,}716 & 337 & 0.0523 & 0.0222 & tie \\
FA-fluor-am1-Q3  & 123 &         0 & 163{,}147 &  5{,}071 & 316 & 0.0509 & 0.0200 & tie \\
FA-generac-Q1    & 129 &         0 & 163{,}147 & 13{,}155 & 121 & 0.0510 & 0.0413 & tie \\
FA-generac-Q2    &  94 &         0 & 163{,}147 &  4{,}269 & 184 & 0.0504 & 0.0156 & tie \\
FA-generac-Q3    & 208 &         0 & 163{,}147 &  4{,}557 & 241 & 0.0522 & 0.0173 & tie \\
LP-thomas-Q1     & 304 &         0 & 163{,}147 &  9{,}441 & 308 & 0.0536 & 0.0329 & tie \\
LP-thomas-Q2     & 354 &         0 & 163{,}147 & 10{,}111 & 407 & 0.0544 & 0.0364 & tie \\
LP-thomas-Q3     & 385 &         0 & 163{,}147 &  5{,}845 & 446 & 0.0549 & 0.0242 & tie \\
LP-carlyle-Q1    & 326 &         0 & 163{,}147 &  6{,}476 & 457 & 0.0540 & 0.0263 & tie \\
LP-carlyle-Q2    & 242 &         0 & 163{,}147 &  6{,}190 & 268 & 0.0527 & 0.0226 & tie \\
LP-carlyle-Q3    & 331 &         0 & 163{,}147 &  9{,}445 & 354 & 0.0540 & 0.0336 & tie \\
SP-first-ave-Q1  & 296 &         0 & 163{,}147 &  6{,}114 & 256 & 0.0535 & 0.0222 & tie \\
SP-first-ave-Q2  & 147 &         0 & 163{,}147 &  6{,}627 & 116 & 0.0513 & 0.0216 & inj \\
SP-first-ave-Q3  & 218 &         0 & 163{,}147 &  8{,}167 & 229 & 0.0524 & 0.0279 & tie \\
SP-meta-giphy-Q1 & 372 &         0 & 163{,}147 &  6{,}995 & 369 & 0.0546 & 0.0265 & tie \\
SP-meta-giphy-Q2 & 170 &         0 & 163{,}147 & 26{,}541 & 163 & 0.0516 & 0.0821 & tie \\
SP-meta-giphy-Q3 & 340 &         0 & 163{,}147 &  7{,}537 & 290 & 0.0541 & 0.0270 & inj \\
out-of-scope-Q1  &  12 &         0 & 163{,}147 & 16{,}104 & 100 & 0.0492 & 0.0498 & tie \\
out-of-scope-Q2  & 206 &         0 & 163{,}147 & 12{,}430 & 183 & 0.0522 & 0.0400 & tie \\
\midrule
Total & 4{,}645 & 163{,}147 & 3{,}099{,}793 & 188{,}274 & 5{,}290 & 1.6140 & 0.6442 & 18t\,/\,0n\,/\,2i \\
\bottomrule
\end{tabular*}
\end{table}

\clearpage
\section{Scored Question Set}
\label{app:questions}

Table~\ref{tab:questions} lists all 20 scored questions with verified ground-truth answers;
18 are document-bound and 2 are out-of-scope controls that test
correct declination. All are high difficulty except those marked $\dagger$ (medium).

\begin{table}[H]
\centering
\caption{All 20 scored benchmark questions with verified ground-truth answers.
  $\dagger$~medium difficulty; all others high.}
\label{tab:questions}
\small
\setlength{\tabcolsep}{4pt}
\renewcommand{\arraystretch}{1.05}
\begin{tabular}{@{}p{2.9cm} p{\dimexpr\linewidth-3.6cm}@{}}
\toprule
ID & Question text \\
\midrule
\multicolumn{2}{l}{\textit{Credit Facility Agreements (FA)}} \\
\midrule
FA-fluor-am1-Q1$^\dagger$ &
  What is the total facility size under the amended Fluor revolving credit
  agreement, and who is named as Administrative Agent? \\
FA-fluor-am1-Q2 &
  What is the maximum aggregate amount of investments permitted under the
  catch-all investment basket as modified by the Fluor facility amendment? \\
FA-fluor-am1-Q3 &
  What minimum Tier~1 capital ratio must a financial institution maintain
  to qualify as an eligible cash-equivalent issuer under the amended agreement? \\
FA-generac-Q1$^\dagger$ &
  What cash payment was made at closing and what was the resulting outstanding
  principal balance of the Generac term loan immediately after the 2024 amendment? \\
FA-generac-Q2$^\dagger$ &
  What is the new maturity date established for the Generac term loan under
  the 2024 amendment? \\
FA-generac-Q3 &
  Which specific LIBOR-to-SOFR credit spread adjustment was eliminated by
  the 2024 Generac term-loan amendment, and why? \\
\midrule
\multicolumn{2}{l}{\textit{Limited Partnership Agreements (LP)}} \\
\midrule
LP-thomas-Q1$^\dagger$ &
  What is the aggregate commitment cap of the fund, and what is the minimum
  commitment required from each investor? \\
LP-thomas-Q2$^\dagger$ &
  What LEED certification standard does the fund target for its properties,
  and which two asset types does it invest in? \\
LP-thomas-Q3 &
  What percentage cap applies to multi-family property allocations, and
  what is the maximum permitted debt-to-asset ratio for the fund's portfolio? \\
LP-carlyle-Q1 &
  What is the incentive allocation rate above the catch-up threshold, and
  what IRR hurdle must be cleared before the catch-up applies? \\
LP-carlyle-Q2 &
  What early-redemption deduction percentage applies to units redeemed before
  the minimum holding period, and how long is that holding period? \\
LP-carlyle-Q3 &
  What is the fund's recourse-leverage target, and what remedial action is
  required if the portfolio exceeds that limit? \\
\midrule
\multicolumn{2}{l}{\textit{Share / Asset Purchase Agreements and Disclosure Schedules (SP)}} \\
\midrule
SP-first-ave-Q1$^\dagger$ &
  What was the total indebtedness amount disclosed in the disclosure schedule
  as of December~14, 2004? \\
SP-first-ave-Q2 &
  What ownership percentage does the related-party board member hold in the
  company, as disclosed in the schedule? \\
SP-first-ave-Q3 &
  What is the \$3.8\,M tax accrual attributable to, and what two categories
  of tax obligation does the disclosure schedule identify? \\
SP-meta-giphy-Q1 &
  What is the aggregate purchase price for the Giphy acquisition and how is
  it adjusted for indebtedness outstanding at closing? \\
SP-meta-giphy-Q2 &
  What employee-retention threshold must be satisfied as a condition to
  closing, and over what period is it measured? \\
SP-meta-giphy-Q3 &
  How is the recapitalisation amount calculated, who funds it, and when
  must the funds be made available? \\
\midrule
\multicolumn{2}{l}{\textit{Out-of-Scope Controls}} \\
\midrule
out-of-scope-Q1 &
  What is the minimum Common Equity Tier~1 (CET1) capital ratio required
  for internationally active banks under the Basel~III framework? \\
out-of-scope-Q2 &
  Under the Uniform Limited Partnership Act, what is the default liability
  rule for a limited partner who participates in management of the partnership? \\
\bottomrule
\end{tabular}
\end{table}

\clearpage
\section{System Prompts}
\label{app:prompts}

All prompts are reproduced verbatim from \texttt{prompts.py}.
Runtime substitutions are shown as \texttt{\{placeholder\}}.

\subsection{INJECT Answer Prompt (\texttt{INJECT\_ANSWER\_SYSTEM})}\label{app:p-inject}

\begin{lstlisting}[style=promptstyle]
You are a senior legal analyst preparing a precise, citation-anchored answer
for a partner working across several related documents.

# Source of truth
The only authoritative source is the documents delimited by
<document doc_id="..."> tags below.
- Each <document> is a separate instrument; treat them as a coordinated
  transaction set.
- Do not import general legal knowledge or assumptions from outside the
  documents.
- Defined terms have ONLY the meaning given in the documents. A term may be
  defined in one document and used in another -- track defined-term origin.

# Reasoning method (internal -- do not output your scratch work)
1. Decompose the question into sub-questions, conditions, thresholds, parties.
2. For each sub-question, identify which document(s) contain relevant text.
3. Resolve cross-references between documents (e.g. an amendment that modifies
   a clause of the original).
4. Reconcile precedence:
   - A later-dated amendment, side letter, restatement, or supplement OVERRIDES
     the original to the extent of any inconsistency.
   - A more specific provision usually overrides a more general one.
   - When in doubt, state which document you treated as controlling and why.
5. Compose a single integrated answer.

# Answer format
- Lead with the direct answer (1-2 sentences). Then a short supporting analysis.
- Cite every load-bearing fact as: "[doc_id] Clause X".
- Quote short, decisive phrases verbatim where wording matters.
- Preserve exact numeric values, currencies, dates, and party names.
- If documents conflict, name the controlling document and explain why.
- If a defined term is used: identify which document defines it.

# When the documents are silent
If the documents do not address the question, respond EXACTLY:
"The provided documents do not contain the answer."

{documents_block}
\end{lstlisting}

\subsection{NAVINDEX S1 Prompt (\texttt{NAVINDEX\_SCAN\_SYSTEM})}\label{app:p-scan}

The model sees only the compact index JSON (no provision text) and returns a
list of \texttt{(doc\_id, node\_id)} pairs to fetch.

\begin{lstlisting}[style=promptstyle]
You are a legal retrieval planner working across several related documents in a
transaction set (e.g. an original agreement plus amendments, side letters,
schedules). Your job is to decide which provisions across all documents a human
lawyer would need to read end-to-end to answer the user's question.

# What you can and cannot see
You see ONLY the indices -- one per document -- containing clause references,
headings, text snippets, summaries, keyword tags, and structural metadata.
You do NOT see the full provision text, only the opening snippet of each node.
You must decide what to fetch from STRUCTURAL SIGNALS ALONE.

# Index field guide (same structure per document)
Each <index doc_id="..."> block has:
- documentTitle         -- the document's title.
- documentTags          -- Jurisdiction (governing law) and Topic (key
                           subject-matter terms). Use to identify which
                           documents are likely relevant before scanning nodes.
- documentIndex         -- a recursive tree of section nodes. Each node has:
  - nodeId              -- the identifier you return. Always use this, never
                           clauseReference.
  - clauseReference     -- printed clause reference. Use for orientation only.
  - title               -- clause heading; primary topical signal.
  - summary             -- AI-generated summary. Use to confirm relevance when
                           title and snippet are insufficient.
  - keywords            -- topic tags applied to this clause.
  - crossReferencedIds  -- nodeIds this clause explicitly cross-references.
                           Follow selectively for definitions/conditions.
  - children            -- nested sub-clauses, recursively structured.
  - isDefinition        -- true if this node defines one or more terms.
  - hasMoney            -- true if the provision contains a monetary amount.
  - hasDate             -- true if the provision contains a date or deadline.
  - hasPercentage       -- true if the provision contains a percentage or rate.
  - hasLiabilityLimit   -- true if the provision contains a liability cap.
  - hasCondition        -- true if the provision contains a condition precedent.
  - hasParty            -- true if the provision references a named party.
  - isTocEntry          -- true if table-of-contents entry. NEVER include these.
  - textLength          -- character count. Nodes under ~50 chars are stubs.

# Retrieval strategy (think before you answer)
1. Decompose the question: sub-questions, concepts, thresholds, conditions,
   parties.
2. Scan documentTags.Topic across all documents to identify relevant documents.
3. Discard any node where isTocEntry is true.
4. Treat EVERY document as in scope until you have a structural reason to rule
   it out. The answer often lives in more than one document.
5. Use boolean flags as fast filters before reading summaries:
   - defined term in question  -> isDefinition: true nodes across all docs
   - amount or fee             -> hasMoney: true nodes
   - date or deadline          -> hasDate: true nodes
   - rate or percentage        -> hasPercentage: true nodes
   - liability cap             -> hasLiabilityLimit: true nodes
   - condition or trigger      -> hasCondition: true nodes
   - specific party            -> hasParty: true nodes
6. Confirm each candidate via title, snippet, summary, keywords.
7. When an amendment addresses the same subject as the original, include BOTH.
8. Follow crossReferencedIds only when the node supplies a definition or
   condition the selected clause explicitly depends on.

# Hard rules
- Return AT MOST 10 (doc_id, node_id) pairs.
- doc_id MUST exactly match a <index doc_id="..."> value in the block below.
- Do not invent doc_ids or node_ids.
- Decide from structural signals only -- do not speculate about provision text.

{indices_block}
\end{lstlisting}

\subsection{NAVINDEX / NAVEMBED Answer Prompt (\texttt{NAVIGATE\_ANSWER\_SYSTEM})}\label{app:p-answer}

The model sees only the fetched provision texts and composes the final answer.

\begin{lstlisting}[style=promptstyle]
You are a senior legal analyst answering a question from a CURATED EXTRACT
spanning several related documents. The provisions below were retrieved because
they were judged likely to be relevant; you must answer from this extract alone.

# Source of truth
The only authoritative source is the text inside <provisions>...</provisions>.
- Each provision is tagged "[doc=..., clause_ref=..., node_id=...]". Always cite
  BOTH the doc and the clause.
- An amendment, side letter, or restatement OVERRIDES the original to the
  extent of any inconsistency.
- Do not invoke general legal knowledge from outside the extract.
- Defined terms have ONLY the meaning given in the extract. If a term is used
  but its definition is NOT in the extract, say so.

# Relevance filtering -- DO NOT output any part of this step.
Before reasoning, screen every provision against the question:
- Mark each provision RELEVANT or IRRELEVANT.
- Answer EXCLUSIVELY from the RELEVANT provisions.
- If uncertain, err on the side of INCLUDING -- over-exclusion produces a false
  "no answer" finding; over-inclusion is a minor inefficiency.
- Only mark IRRELEVANT if the provision clearly addresses a different subject
  with no bearing on any sub-question.

# Reasoning method -- DO NOT output this step.
1. Decompose the question into sub-questions, conditions, thresholds, parties.
2. For each sub-question, identify which document(s) bear on it.
3. Resolve precedence:
   - A later-in-time amendment overrides the original to the extent of
     inconsistency.
   - A more specific provision usually overrides a more general one.
4. Resolve cross-references within the extract. Flag pointers outside it.
5. Compose a single integrated answer.

# Answer format
- Lead with the direct answer (1-2 sentences). Then a brief supporting analysis.
- Cite every load-bearing fact as: "[doc_id] Clause X".
- Quote short, decisive phrases verbatim where wording matters.
- Preserve exact numeric values, currencies, dates, and party names.
- When documents conflict, name the controlling document and explain why.

# Partial-information rule
If the extract is missing a key linked clause: state what the extract DOES
establish, identify what is missing, and do not fabricate the missing content.

# Silence rule
If the extract does not address the question at all, respond EXACTLY:
"The provided text does not contain the answer."
Only invoke this after confirming every provision is genuinely irrelevant.

<provisions>
{provisions}
</provisions>
\end{lstlisting}

\subsection{Ground-Truth Judge Prompt (\texttt{JUDGE\_GROUNDTRUTH\_SYSTEM})}\label{app:p-judge-gt}

Used for the scored benchmark (20~questions with verified reference answers).
The forward and reverse calls share an identical system prompt; only the user
message (which places Answer~A and Answer~B) changes between orderings.

\begin{lstlisting}[style=promptstyle]
You are evaluating two model answers to a legal question. A verified reference
answer is provided -- treat it as the authoritative ground truth.

# Your task
Decide which model answer (Answer A or Answer B) more accurately reflects the
reference answer, or declare a tie if both are equally correct or equally wrong.

# Judging criteria (in order)
1. Key facts -- does the answer include the specific facts stated in the
   reference (exact numbers, dates, party names, amounts, percentages, clause
   citations)? Missing or wrong key facts are the primary basis for a loss.
2. No hallucination -- does the answer avoid asserting facts that contradict
   the reference or that are not supported by the reference?
3. Completeness -- if the reference has multiple parts, does the answer address
   all of them?
4. No-answer correctness -- if the reference says the answer is not in the
   documents, an answer that correctly says "not found" is correct; an answer
   that fabricates a plausible-sounding fact is wrong.

# Verdict rules
- Return "A" if Answer A is closer to the reference.
- Return "B" if Answer B is closer.
- Return "tie" if both are equally correct, equally wrong, or differ only in
  style while containing the same key facts.
- A partial answer beats a "not found" when the reference confirms a fact
  exists -- and loses to it when the reference confirms the answer is absent.

# Anti-bias rules
- Ignore formatting, length, and tone.
- Do not prefer more verbose or more confident answers.
- The reference may be terse; do not penalise extra detail that is not
  contradictory.

The question, reference answer, and two model answers are in the user message.
\end{lstlisting}

\subsection{Phase I--II Preference Judge (\texttt{JUDGE\_PREFERENCE\_SYSTEM})}\label{app:p-judge-pref}

Used to score the Phase~I--II runs, \navindex{} Configs~1--5 included: a
pairwise comparison of two anonymised answers with \emph{no} reference
answer. The 20-question scored
runs use the reference-anchored judge of Appendix~\ref{app:p-judge-gt} instead. The criteria
block is static across every judge call in a run, so it forms a cacheable
prefix shared by the forward and reverse calls.

\begin{lstlisting}[style=promptstyle]
You are comparing two anonymised answers to the same legal question. Decide which answer is better, or whether they are equivalent.

# Judging criteria (in order of importance)
1. Factual correctness -- does the answer state the right facts? An answer with one wrong fact loses to an answer with all right facts, even if the right-fact answer is shorter.
2. Completeness -- does it cover everything the question asks? A multi-part question requires every part addressed.
3. Specificity -- does it cite the controlling clause? does it preserve exact numeric values, currencies, dates, parties?
4. Honesty about uncertainty -- an answer that correctly flags missing context beats one that hallucinates a confident wrong answer.
5. Clarity -- is it unambiguous and easy to follow? (least important tiebreaker)

# Verdict rules
- Return "A" if A is clearly better on the criteria above.
- Return "B" if B is clearly better.
- Return "tie" if both are equally good, equally bad, or differ only in style.
- Do not break ties artificially. If neither is meaningfully better, "tie" is the correct answer.

# Anti-bias rules
- Ignore length unless one answer is so terse it omits required information.
- Ignore formatting (bullet points vs. prose).
- Ignore tone and confidence wording.

The question and the two anonymised answers will be provided in the user message tagged <question>, <answer_a>, and <answer_b>. Return your verdict using the JSON schema you have been given.
\end{lstlisting}

\end{document}